\documentclass[runningheads]{llncs}
\usepackage{graphicx}
\usepackage{amsmath,amssymb} % define this before the line numbering.
\usepackage{color}
\usepackage{bm}
\newcommand{\ak}{\textcolor{black}}

\begin{document}
\title{ADVISE: Symbolism and External Knowledge for Decoding Advertisements} 
% Replace with your title

\titlerunning{ADVISE: Symbolism and External Knowledge for Decoding Advertisements}
% Replace with a meaningful short version of your title
%
% \author{First Author\inst{1}\orcidID{0000-1111-2222-3333} \and
% Second Author\inst{2,3}\orcidID{1111-2222-3333-4444} \and
% Third Author\inst{3}\orcidID{2222--3333-4444-5555}}
%\author{Keren Ye\orcidID{0000-0002-7349-7762} \and
%Adriana Kovashka\orcidID{0000-0003-1901-9660}}
\author{Keren Ye \and Adriana Kovashka}
%
%Please write out author names in full in the paper, i.e. full given and family names. 
%If any authors have names that can be parsed into FirstName LastName in multiple ways, please include the correct parsing, in a comment to the volume editors:
%\index{Lastnames, Firstnames}
%(Do not uncomment it, because you may introduce extra index items if you do that, we will use scripts for introducing index entries...)
\authorrunning{Keren Ye \and Adriana Kovashka}
% Replace with shorter version of the author list. If there are more authors than fits a line, please use A. Author et al.
%

% \institute{Princeton University, Princeton NJ 08544, USA \and
% Springer Heidelberg, Tiergartenstr. 17, 69121 Heidelberg, Germany
% \email{lncs@springer.com}\\
% \url{http://www.springer.com/gp/computer-science/lncs} \and
% ABC Institute, Rupert-Karls-University Heidelberg, Heidelberg, Germany\\
% \email{\{abc,lncs\}@uni-heidelberg.de}}

\institute{University of Pittsburgh, Pittsburgh PA 15260, USA\\
\email{\{yekeren,kovashka\}@cs.pitt.edu}}

\maketitle              % typeset the header of the contribution
\begin{abstract}
In order to convey the most content in their limited space, advertisements embed references to outside knowledge via symbolism. For example, a motorcycle stands for adventure (a positive property the ad wants associated with the product being sold), and a gun stands for danger (a negative property to dissuade viewers from undesirable behaviors). We show how to use symbolic references to better understand the meaning of an ad. We further show how anchoring ad understanding in general-purpose object recognition and image captioning improves results. We formulate the ad understanding task as matching the ad image to human-generated statements that describe the action that the ad prompts, and the rationale it provides for taking this action. Our proposed method outperforms the state of the art on this task, and on an alternative formulation of question-answering on ads. We show additional applications of our learned representations for matching ads to slogans, and clustering ads according to their topic, without extra training.

% \keywords{First keyword  \and Second keyword \and Another keyword. We would like to encourage you to list your keywords within the abstract section}
\end{abstract}

\section{Introduction}
\label{sec:intro}

Advertisements are a powerful tool for affecting human behavior. 
Product ads convince us to make large purchases, e.g. for cars and home appliances, or small but recurrent purchases, e.g. for laundry detergent. 
Public service announcements (PSAs) encourage socially beneficial behaviors, e.g. combating domestic violence or driving safely.
To stand out from the rest, ads have to be both eye-catching and memorable \cite{young2005advertising}, while also conveying the information that the ad designer wants to impart. 
All this must be done in a limited space (one image) and time (however many seconds the viewer spends looking at the ad). 

How can ads get the most ``bang for their buck''? 
One technique is to make references to knowledge viewers already have, e.g. cultural knowledge, associations, and \emph{symbolic mappings} humans have learned \cite{scott1994images,levy1959symbols,spears1996symbolic,leigh1992symbolic}. 
These symbolic references might come from literature (e.g. a snake symbolizes evil or danger), movies (a motorcycle symbolizes adventure or coolness), common sense (a flexed arm symbolizes strength), or pop culture (Usain Bolt symbolizes speed). 

In this paper, we describe how to use symbolic mappings to predict the messages of advertisements. 
On one hand, we model how components of the ad image serve as visual anchors to concepts outside the image, using annotations in the Ads Dataset of \cite{Hussain_2017_CVPR}.
On the other hand, we use knowledge sources external to the main task, such as object detection models, to better relate ad images to their corresponding messages.
Both of these are forms of using outside knowledge, and they both boil down to learning links between objects and symbolic concepts.
We use each type of knowledge in two ways, as a constraint or as an additive component for the learned image representation. 

\begin{figure}[t]
    \centering
    \includegraphics[width=1\linewidth]{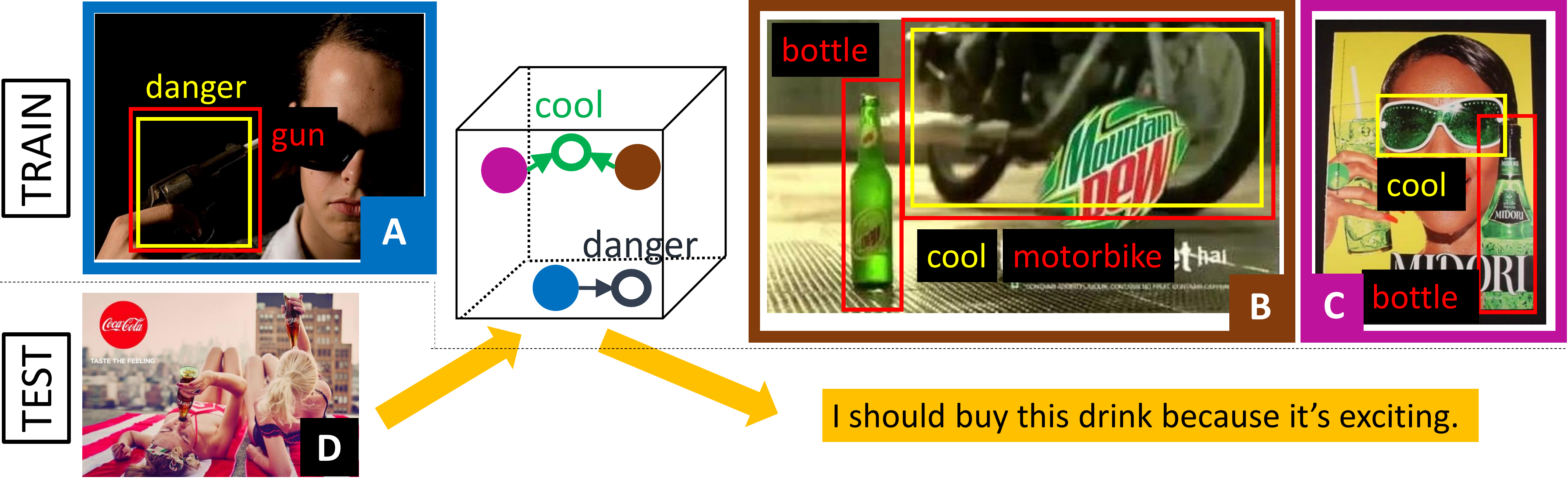}
    \caption{Our key idea: Use symbolic associations shown in yellow (a gun symbolizes danger; a motorcycle symbolizes coolness) and recognized objects shown in red, to learn an image-text space where each ad maps to the correct statement that describes the message of the ad. The symbol ``cool'' brings images B and C closer together in the learned space, and further from image A and its associated symbol ``danger.'' At test time (shown in orange), we use the learned image-text space to retrieve a matching statement for test image D. At test time, the symbol labels are \emph{not} provided.}
    \label{fig:concept}
\end{figure}

We focus on the following multiple-choice task, implemented via ranking: Given an image and several statements, the system must identify the correct statement to pair with the ad. For example, for test image D in Fig.~\ref{fig:concept}, the system might predict the right statement is ``Buy this drink because it's exciting.''
Our method learns a joint image-text embedding that associates ads with their corresponding messages. 
The method has three components: (1) an image embedding which takes into account individual regions in the image, (2) constraints on the learned space from symbol labels and object predictions, and (3) an additive expansion of the image representation using a symbol distribution.  
\ak{These three components are shown in Fig.~\ref{fig:concept},
and all of them rely on  external knowledge in the form of symbols and object predictions.}
Note that we can recognize the symbolic association to danger in Fig.~\ref{fig:concept} via two channels: either a direct classifier that learns to link certain visuals to the ``danger'' concept, or learning associations between actual \emph{objects} in the image which can be recognized by object detection methods (e.g. ``gun''), and symbolic concepts. 
We call our method ADVISE: \textbf{AD}s \textbf{VI}sual \textbf{S}emantic \textbf{E}mbedding.

We primarily focus on public service announcements, rather than product (commercial) ads. PSAs tend to be more conceptual and challenging, often involving multiple steps of reasoning. Quantitatively, 59\% of the product ads in the dataset of \cite{Hussain_2017_CVPR} are straightforward, i.e. would be nearly solved with traditional recognition advancements. In contrast, only 33\% of PSAs use straightforward strategies, while the remaining 67\% use challenging non-literal rhetoric. 
Our method outperforms several recent baselines, including prior visual-semantic embeddings \cite{faghri2017vse++,Eisenschtat_2017_CVPR} and methods for understanding ads \cite{Hussain_2017_CVPR}. 

In addition to showing how to use external knowledge to solve ad-understanding, we demonstrate how recent advances in object recognition help with this task. 
While \cite{Hussain_2017_CVPR} evaluates basic techniques, it does not employ recent advances like region proposals \cite{He_2017_ICCV,ren2015faster,liu2016ssd,girshick2014rich} or attention \cite{Chen_2016_CVPR,You_2016_CVPR,Xu2016,Shih_2016_CVPR,Yang_2016_CVPR,Ren_2017_CVPR,nam2017dual,lu2017knowing,fu2017look,Zheng_2017_ICCV,Pedersoli_2017_ICCV}.
%, or image-text embeddings \cite{faghri2017vse++,kiros2014unifying,Cao_2017_CVPR,Chen_2017_CVPR,Eisenschtat_2017_CVPR,Gomez_2017_CVPR,Plummer_2017_CVPR}.

To summarize, our contributions are as follows:
% \begin{itemize}[itemsep=2pt,topsep=2pt,parsep=1pt,partopsep=1pt]
\begin{itemize}
    \item We show how to effectively use symbolism to better understand ads.  
    \item We show how to make use of noisy caption predictions to bridge the gap between the abstract task of predicting the message of an ad, and more accessible information such as the objects present in the image. Detected objects are mapped to symbols via a domain-specific knowledge base.
    \item We improve the state of the art in understanding ads by 21\%.% in the case of public service announcements, and 25\% in the case of product ads.
    \item We show for ``abstract'' PSAs, conceptual knowledge helps more, while for product ads, general-purpose object recognition techniques are more helpful. 
\end{itemize}

The remainder of the paper is organized as follows. We overview related work in Sec.~\ref{sec:related}. In Sec.~\ref{sec:task}, we describe our ranking task, and in Sec.~\ref{sec:triplet}, we describe standard triplet embedding on ads. In Sec.~\ref{sec:regions}, we discuss the representation of an image as a combination of region representations, weighed by their importance via an attention model. In Sec.~\ref{sec:symbols}, we describe how we use external knowledge to constrain the learned space. In Sec.~\ref{sec:knowledge_base}, we develop an optional additive refinement of the image representation.
%, again using external knowledge and symbols. 
In Sec.~\ref{sec:results}, we compare our method to the state of the art, and conduct ablation studies. We conclude in Sec.~\ref{sec:conclusion}.

\section{Related Work}
\label{sec:related}

\paragraph{Advertisements and multimedia.}

The most related work to ours is \cite{Hussain_2017_CVPR} which proposes the problem of decoding ads, formulated as answering the question ``\emph{Why} should I [action]?'' where [action] is what the ad suggests the viewer should do, e.g. buy a car or help prevent domestic violence. The dataset contains 64,832 image ads. Annotations include the topic (product or subject) of the ad, sentiments and actions the ad prompts, rationales provided for why the action should be done, symbolic mappings (referred to as signifier-signified, e.g. motorcycle-adventure), etc. 
Considering the media domain more broadly, \cite{joo2014visual} analyze in what light a photograph portrays a politician, and \cite{joo2015automated} examine how the facial features of a candidate determine the outcome of an election. 
This work only applies to images of people.
Also related is work in parsing infographics, charts and comics \cite{bylinskii2017understanding,kembhavi2016diagram,Iyyer_2017_CVPR}.
In contrast to these, our interest is analyzing the \emph{implicit} arguments ads were created to make. 

\paragraph{Vision, language and image-text embeddings.}

Recently there is great interest in joint vision-language tasks, e.g. captioning \cite{vinyals2015show,Karpathy_2015_CVPR,Donahue_2015_CVPR,Johnson_2016_CVPR,Hendricks_2016_CVPR,You_2016_CVPR,Venugopalan_2017_CVPR,Vedantam_2017_CVPR,Yu_2017_CVPR_gaze,Yang_2017_CVPR,Gan_2017_CVPR,Pedersoli_2017_ICCV,Shetty_2017_ICCV,show_adapt_tell,Krishna_2017_ICCV}, visual question answering \cite{Antol_2015_ICCV,Yu_2015_ICCV,Malinowski_2015_ICCV,Yang_2016_CVPR,Shih_2016_CVPR,Wu_2016_CVPR,Tapaswi_2016_CVPR,Zhu_2016_CVPR,Zhu_2017_CVPR,Hu_2017_ICCV,wang2017fvqa,johnson2017inferring,Teney_2018_CVPR}, and cross-domain retrieval \cite{Chen_2017_CVPR,Cao_2017_CVPR,Yu_2017_CVPR_text,Li_2017_ICCV}. 
These often rely on learned image-text embeddings. 
\cite{faghri2017vse++,kiros2014unifying} use triplet loss where an image and its corresponding human-provided caption should be closer in the space than pairs that do not match.
\cite{Eisenschtat_2017_CVPR} propose a bi-directional network to maximize correlation between matching images and text, akin to CCA \cite{hotelling1936relations}.
None of these consider images with implicit persuasive intent, as we do. 
We compare against \cite{faghri2017vse++,Eisenschtat_2017_CVPR} in Sec.~\ref{sec:results}.
%\cite{Gomez_2017_CVPR} utilize the images and texts in Wikipedia articles for self-supervision, where an image representation is learned that predicts the same topics that can be extracted from the text.
%\cite{Cao_2017_CVPR} learn a quantization in the transformed space of CNN features such that the similarity between reconstructed images and the correct textual description is preserved, in order to enable efficient image retrieval.
%\cite{Chen_2017_CVPR} improve retrieval results through attention modules that operate within the image and text modalities alone, as well as one linking the two modalities.
%\cite{Plummer_2017_CVPR} learn a visual-semantic embedding to allow text-guided summarization of videos.

\paragraph{External knowledge for vision-language tasks.}

%We propose to use external knowledge for decoding ads, in two ways: via symbols that inherently refer to outside knowledge, and by using outside knowledge to learn to detect symbols. 
\cite{Wu_2016_CVPR,wang2017fvqa,johnson2017inferring,Zhu_2017_CVPR,Teney_2018_CVPR} examine the use of knowledge bases and perform explicit reasoning for answering visual questions.
\cite{Venugopalan_2017_CVPR} use external sources to diversify their image captioning model.
\cite{Misra_2017_CVPR} learn to compose object classifiers by relating semantic and visual similarity. 
\cite{Marino_2017_CVPR,goo2016taxonomy} use knowledge graphs or hierarchies to aid in object recognition. 
These works all use mappings that are objectively/scientifically grounded, i.e. lion is a type of cat.
In contrast, we use cultural associations that arose in the media/literature and are internalized by humans, e.g. motorcycles are associated with adventure.

\paragraph{Region proposals and attention.}

Region proposals \cite{He_2017_ICCV,ren2015faster,liu2016ssd,girshick2014rich} guide an object detector to regions likely to contain objects. 
Attention \cite{Chen_2016_CVPR,You_2016_CVPR,Xu2016,Shih_2016_CVPR,Yang_2016_CVPR,Ren_2017_CVPR,nam2017dual,lu2017knowing,fu2017look,Zheng_2017_ICCV,Pedersoli_2017_ICCV}  focuses prediction tasks on regions likely to be relevant. 
We show that for our task, the attended-to regions must be those likely to be visual anchors for symbolic references.

\section{Approach}
\label{sec:approach}

We learn an embedding space where we can evaluate the similarity between ad images and ad messages.
We use symbols and external knowledge in three ways: by representing the image as a weighted average of its regions that are likely to make symbolic references (Sec.~\ref{sec:regions}), by enforcing that images with the same symbol labels or detected objects  are close (Sec.~\ref{sec:symbols}), and by enhancing the image representation via an attention-masked symbol distribution (Sec.~\ref{sec:knowledge_base}).
In Sec.~\ref{sec:results} we demonstrate the utility of each component.

\subsection{Task and dataset}
\label{sec:task}

%We develop a method for understanding advertisements. We consider the task of ranking statements that capture the rhetoric of ads in the dataset of  \cite{Hussain_2017_CVPR}.

In \cite{Hussain_2017_CVPR}, the authors tackled  answering the question ``Q: Why should I [action]?'' with ``A: [one-word reason].'' An example question-answer pair is ``Q: Why should I speak up about domestic violence? A: bad.'' In other words, question-answering is formulated as a classification task.
The ground-truth one-word answers in \cite{Hussain_2017_CVPR}'s evaluation are picked from human-provided full-sentence answers, also available in the dataset.
However, using a single word is insufficient to capture the rhetoric of complex ads. On one hand, summarizing the full sentence using only one word is too challenging, for example, for the question ``Q: Why should I buy authentic Adidas shoes?'', the ground-truth answer ``feet'' used in \cite{Hussain_2017_CVPR} cannot convey both the meaning of ``protect'' and ``feet'' while the full-sentence answer ``Because it will protect my feet'' does capture both. On the other hand, picking one word as the answer may be misleading and imprecise, for example, for the ``Q: Why should I buy the Triple Double Crunchwrap?'', picking ``short'' from the sentence ``Because it looks tasty and is only available for a short time'' is problematic. Thus, while we show that we outperform prior art on the original question-answering task of \cite{Hussain_2017_CVPR}, we focus on an alternative formulation. 

We ask the system to pick which \emph{action-reason statement} is most appropriate for the image. We retrieve statements in the format: ``I should [action] because [reason].'' e.g. ``I should speak up about domestic violence because \emph{being quiet is as bad as committing violence yourself.}''
For each image, we use three related statements (i.e. statements provided by humans for this image) and randomly sample 47 unrelated statements (written for \emph{other} images). 
The system must rank these 50 statements based on their similarity to the image.

This ranking task is akin to multiple-choice question-answering, which was also used in prior VQA works \cite{Antol_2015_ICCV,Tapaswi_2016_CVPR}, but unlike these, we do not take the question as input. 
Similarly, in image captioning, \cite{Karpathy_2015_CVPR,faghri2017vse++} look for the most suitable image description from a much larger candidates pool.

\subsection{Basic image-text triplet embedding}
\label{sec:triplet}

We first directly learn an embedding that optimizes for the ranking task. 
We require that the distance between an image and its corresponding statement should be smaller than the distance between that image and any other statement, or between other images and that statement. In other words, we minimize:
\begin{equation}
\label{eq:triplet}
\begin{split}
L(\bm{v}, \bm{t}; \bm{\theta})=\sum_{i=1}^K \Big[&\underbrace{\sum_{j \in N_{vt}(i)} \left[ \lVert \bm{v}_i - \bm{t}_i \rVert_2^2 - \lVert \bm{v}_i - \bm{t}_j \rVert_2^2 + \beta \right]_+}_{\text{image as anchor, rank statements}}\\
+ &\underbrace{\sum_{j \in N_{tv}(i)} \left[ \lVert \bm{t}_i - \bm{v}_i \rVert_2^2 - \lVert \bm{t}_i - \bm{v}_j \rVert_2^2 + \beta \right]_+}_{\text{statement as anchor, rank images}}
\Big]
\end{split}
\end{equation}
where $K$ is the batch size; $\beta$ is the margin of triplet loss; $\bm{v}$ and $\bm{t}$ are the visual and textual embeddings we are learning, respectively; $\bm{v}_i$, $\bm{t}_i$ correspond to the same ad; $N_{vt}(i)$ is the negative statement set for the $i$-th image, and $N_{tv}(i)$ is the negative image set for the  $i$-th statement, defined in Eq.~\ref{eq:mining}. \ak{These two negative sample sets involve the most challenging $k'$ examples within the size-$K$ batch. A natural explanation of Eq.~\ref{eq:mining} is that it seeks to find a subset $A\subseteq\{1,...,K\}$ which involves the  $k'$ most confusing examples.}
\begin{equation}
\label{eq:mining}
\begin{split}
N_{vt}(i)=\operatorname*{arg\,min}_{\substack{A\subseteq\{1,...,K\},\\|A|=k'}} \sum_{\substack{j \in A, \\i\neq j}} \lVert \bm{v}_i - \bm{t}_j \rVert_2^2, ~~~~~
N_{tv}(i)=\operatorname*{arg\,min}_{\substack{A\subseteq\{1,...,K\},\\|A|=k'}} \sum_{\substack{j \in A, \\i\neq j}} \lVert \bm{t}_i - \bm{v}_j \rVert_2^2
\end{split}
\end{equation}

\emph{Image embedding.}
We extract the image's Inception-v4 CNN feature (1536-D) using \cite{inceptionv4}, then use a fully-connected layer with parameter $\bm{w} \in \mathbb{R}^{200\times 1536}$ to project it to the 200-D joint embedding space:
\begin{equation} \label{eq:image_embedding}
\bm{v}=\bm{w} \cdot CNN(\bm{x})
\end{equation}

\emph{Text embedding.}
\ak{We use mean-pooling to aggregate word embedding vectors into 200-D text embedding $\bm{t}$ and use GloVe \cite{pennington2014glove} to initialize the embedding matrix. There are two reasons for us to choose mean-pooling: (1) comparable performance to the LSTM\footnote{Non-weighted/weighted mean-pooling of word embeddings achieved 2.45/2.47 rank. The last hidden layer of an LSTM achieved 2.74 rank, while non-weighted/weighted averaging of the hidden layers achieved 2.43/2.46, respectively. Lower is better.}, and (2) better interpretability. By using mean-pooling, image and words are projected to the same feature space, allowing us to assign word-level semantics to an image, or even to image regions. In contrast, LSTMs encode meaning of nearby words which is undesirable for interpretability.}

\emph{\ak{Hard negative mining.}}
Different ads might convey similar arguments, so the sampled negative may be a viable positive. For example, for a car ad with associated statement ``I should buy the car because it's fast'', a hard negative ``I should drive the car because of its speed'' may also be proper.
Using the $k'$ most challenging examples in the size-$K$ batch  (Eq.~\ref{eq:mining}) is our trade-off between using all and using only the most challenging example, inspired by \cite{schroff2015facenet,hermans2017defense,faghri2017vse++,Wu_2017_ICCV}. 
Our experiment (in supp) shows this trade-off is better than either extreme.

\subsection{Image embedding using symbol regions}
\label{sec:regions}

Since ads are carefully designed, they may involve complex narratives with several distinct components, i.e. several regions in the ad might need to be interpreted individually first to decode the full ad's meaning. 
Thus, we represent an image as a collection of its constituent regions, using an attention module to \ak{aggregate all the representations from different regions.}

Importantly, 
the chosen regions should be those likely to serve as visual anchors for symbolic references (such as the motorcycle or shades in Fig.\ref{fig:concept}, rather than the bottles). 
\ak{Thus we consider all the 13,938 images, which are annotated as containing symbols, each with up to five bounding box annotations.}
Our intuition is that ads draw the viewer's attention in a particular way, and the symbol bounding boxes, without symbol labels, can be used to approximate this. 
% This label-agnostic method is a new use of \cite{Hussain_2017_CVPR}'s symbolism data. that has not been explored before. 
More specifically, we use the SSD object detection model \cite{liu2016ssd} implemented by \cite{Huang_2017_CVPR}, pre-train it on the COCO \cite{lin2014microsoft} dataset, and fine-tune it with the symbol bounding box annotations \cite{Hussain_2017_CVPR}. 
We show in Sec.~\ref{sec:quant} that this fine-tuning is crucial, i.e. general-purpose regions such as COCO boxes produce inferior results. 

\begin{figure}[t]
    \centering
    \includegraphics[width=1.0\linewidth]{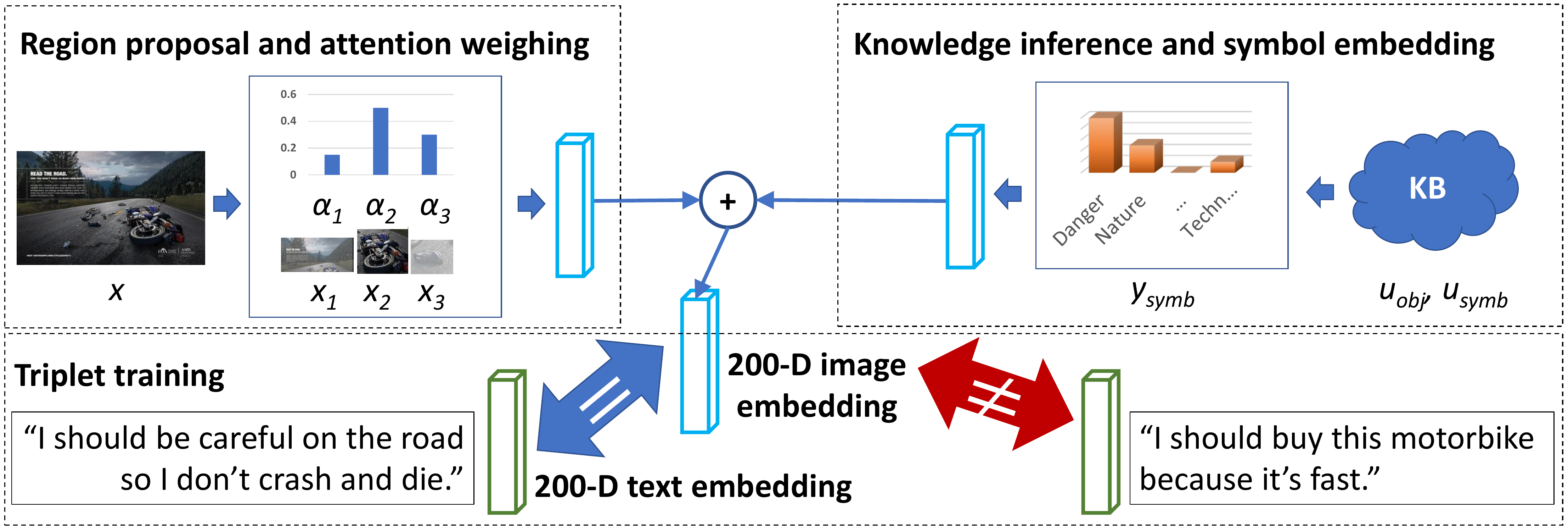}
    \caption{Our image embedding model with knowledge branch. In the main branch (top left), multiple image symbolic anchors are proposed. Attention weighting is applied, and the image is represented as a weighted combination of the regions. The knowledge branch (top right) predicts the existence of symbols, maps these to  200-D, and adds them to the image embedding. We then perform triplet training to learn such an embedding space that keeps images close to their matching action-reason statements.}
    \label{fig:method}
\end{figure}

\ak{We use bottom-up attention \cite{Anderson_2018_CVPR,Teney_2018_CVPR,Krause_2017_CVPR} to aggregate the information from symbolic regions (see Fig.~\ref{fig:method}).
More specifically, we use the Inception-v4 model \cite{inceptionv4} to extract the 1536-D CNN features for all symbol proposals. Then, for each CNN feature $\bm{x}_i, i \in \{1, \dots, M\}$ (we set $M=10$, i.e., 10 proposals per image), a fully-connected layer is applied to project it to}: 1) a 200-D embedding vector $\bm{v}_i$ (Eq.~\ref{eq:region_embedding}, $\bm{w}\in \mathbb{R}^{200\times 1536}$), and 2) a confidence score $a_i$ saying how much  the region should contribute to the final representation (Eq.~\ref{eq:region_attention}, $\bm{w}_a\in \mathbb{R}^{1\times 1536}$). The final image representation $\bm{z}$ is a weighted sum of these region-based vectors (Eq.~\ref{eq:attention_image_embedding}). 

\begin{equation} \label{eq:region_embedding}
\bm{v}_i=\bm{w} \cdot CNN(\bm{x}_i)
\end{equation}
\begin{equation} 
\label{eq:region_attention}
a_i=\bm{w}_{a} \cdot CNN(\bm{x}_i),\quad
\bm{\alpha}=softmax(\bm{a})
\end{equation}
\begin{equation} 
\label{eq:attention_image_embedding}
\bm{z}=\sum\nolimits_{i=1}^{M}\alpha_i \bm{v}_i
\end{equation}

The loss used to learn the image-text embedding is the same as in Eq.~\ref{eq:triplet}, but defined using the region-based image representation $\bm{z}$ instead of $\bm{v}$: $L(\bm{z}, \bm{t}; \bm{\theta})$.

%We demonstrate that (1) learning a region proposal network with attention, and (2) learning from symbol bounding boxes, greatly help the statement retrieval task. In particular, statement ranking results are worse if we use a generic pre-trained region proposal network. We argue that general-purpose object detection models cannot capture nuance in ads since they ignore uncommon or abstract objects. % such as logo, ads text, etc.

\subsection{Constraints via symbols and captions}
\label{sec:symbols}

\ak{We next exploit the symbol labels which are part of \cite{Hussain_2017_CVPR}. Symbols are abstract words such as ``freedom'' and ``happiness'' that provide additional information humans sense from the ads. We add additional constraints to the loss terms such that two images/statements that were annotated with the same symbol are closer in the learned space than images/statements annotated with different symbols}. In the \emph{extra} loss term (Eq.~\ref{eq:triplet_symbols}), $\bm{s}$ is the 200-D embedding of a symbol word; $\bm{z}$ is the 200-D region-based image representation defined in Eq.~\ref{eq:attention_image_embedding}; and $N_{sz}(i)$ and $N_{st}(i)$ are the negative image/statement sets of the $i$-th symbol in the batch, defined similar to Eq.~\ref{eq:mining}.
\begin{equation} 
\label{eq:triplet_symbols}
\begin{split}
L_{sym}(\bm{s}, \bm{z}, \bm{t}; \bm{\theta})=\sum_{i=1}^K \Big[
&\underbrace{\sum_{j \in N_{sz}(i)} \left[ \lVert \bm{s}_i - \bm{z}_i \rVert_2^2 - \lVert \bm{s}_i - \bm{z}_j \rVert_2^2 + \beta \right]_+}_{\text{symbol as anchor, rank images}}\\
+&\underbrace{\sum_{j \in N_{st}(i)} \left[ \lVert \bm{s}_i - \bm{t}_i \rVert_2^2 - \lVert \bm{s}_i - \bm{t}_j \rVert_2^2 + \beta \right]_+}_{\text{symbol as anchor, rank statements}}
\Big]
\end{split}
\end{equation}

% These new constraints allow the model to converge faster and make training more stable. They help create groupings in the learned space containing images with similar symbolism. We explicitly embed symbol labels in the same feature space as images and statements; these symbol embedding vectors serve as entry points for external knowledge and shall be further discussed in Sec.~\ref{sec:knowledge_base}. 

Much like symbols, the objects found in an image are quite telling of the message of the ad. For example, environment ads often feature animals, safe driving ads feature cars, beauty ads feature faces, drink ads feature bottles, etc.
However, since the Ads Dataset contains insufficient data to properly model object categories, we use DenseCap \cite{Johnson_2016_CVPR} to bridge the objects defined in Visual Genome \cite{krishna2017visual} to the ads reasoning statements.
\ak{More specifically, we use the DenseCap model to generate image captions and treat these as pre-fetched knowledge. For example, the caption ``woman wearing a black dress'' provides extra information about the objects in the image: ``woman'' and ``black dress''.}
We create additional constraints: If two images/statements have similar DenseCap predicted captions, they should be closer than images/statements with different captions. The \emph{extra} loss term is defined similar to Eq.~\ref{eq:triplet_symbols}
using $\bm{c}$ for the caption representations.

\ak{In our setting, word embedding weights are not shared among the three vocabularies (ads statement, symbols, and DenseCap predictions). Our consideration is that the meaning of the same surface words may vary in these domains thus they need to have different embeddings. We weigh the symbol-based and object-based constraints by 0.1 since they in isolation do not tell the full story of the ad. We found that it is not sufficient to use \emph{any} type of label as constraint in the domain of interest (see supp): using symbols as constraints gives greater benefit than the topic (product) labels in \cite{Hussain_2017_CVPR}'s dataset, and this point is not discussed in the general proxy learning literature \cite{movshovitz2017no}.}

\subsection{Additive external knowledge} \label{sec:knowledge_base}

In this section, we describe how to make use of external knowledge that is adaptively added, to compensate for inadequacies of the image embedding. This external knowledge can take the form of a mapping between physical objects and implicit concepts, or a classifier mapping pixels to concepts. 
Given a challenging ad, a human might look for visual cues and check if they remind him/her of concepts (e.g. ``danger'', ``beauty'', ``nature'') seen in other ads. 
Our model interprets ads in the same way: based on an external knowledge base, it \emph{infers} the abstract symbols. 
In contrast to Sec.~\ref{sec:symbols} which uses the \emph{annotated} symbols at training time, here we use a \emph{predicted} symbol distribution at both training and test time as a secondary image representation.
Fig.~\ref{fig:method} (top right) shows the general idea of the external knowledge branch. Note our model only uses external knowledge to compensate its own lack of knowledge \ak{(since we train the knowledge branch after the convergence of the visual semantic embedding branch)}, and it assigns small weights for uninformative \ak{knowledge}.

We propose two ways to additively expand the image representation with external knowledge, and describe \emph{two ways of setting} $\bm{y}_{symb}$ in  Eq.~\ref{eq:final}. Both ways are a form of knowledge base (KB) mapping physical evidence to concepts. 

\emph{\textbf{KB Symbols.}} The first way is to directly train classifiers to link certain visuals to symbolic concepts. We learn a multilabel classifier $\bm{u}_{symb}$ to obtain a symbol distribution $\bm{y}_{symb} = \textit{sigmoid}(\bm{u}_{symb} \cdot \bm{x})$. We learn a weight $\alpha^{symb}_j$ for each of $j \in \{1, \dots, C=53\}$ symbols from the Ads Dataset, denoting 
whether a particular symbol is helpful for the statement matching task.

\emph{\textbf{KB Objects.}}
The second method is to learn associations between surface words for detected objects and abstract concepts. 
For example, what type of ad might I see a ``car'' in? What about a ``rock'' or ``animal''? 
We first construct a knowledge base associating object words to symbol words. 
We compute the similarity in the learned image-text embedding space between symbol words and DenseCap words, then create a mapping rule (``[object] implies [symbol]'') for each symbol and its five most similar DenseCap words. This results in a 53$\times V$ matrix $\bm{u}_{obj}$, where $V$ is the size of DenseCap's vocabulary. Each row contains five entries of 1 denoting the mapping rule, and $V-5$ entries of 0. 
Examples of learned mappings are shown in Table \ref{tab:synonyms}. 
For a given image, we use \cite{Johnson_2016_CVPR} to predict the three most probable words in the DenseCap vocabulary, and put the results in a multi-hot  $\bm{y}_{obj} \in \mathbb{R}^{V \times 1}$ vector. We then matrix-multiply to accumulate evidence for the presence of all symbols using the detected objects: $\bm{y}_{symb} = \bm{u}_{obj} \cdot \bm{y}_{obj}$. 
We associate a weight $\alpha^{symb}_{jl}$ with each rule in the KB.

For both methods, we first use the attention weights $\alpha^{symb}$ as a mask, then project the 53-D symbol distribution $\bm{y}_{symb}$ into 200-D, and add it to the image embedding.
This additive branch is most helpful when the information it contains is not already contained in the main image embedding branch. We found this happens when the discovered symbols are rare. 

\subsection{ADVISE: our final model}
\label{sec:advise_final}

Our final \textbf{AD}s \textbf{VI}sual \textbf{S}emantic \textbf{E}mbedding loss combines the losses from Sec.~\ref{sec:triplet}, \ref{sec:regions}, \ref{sec:symbols}, and \ref{sec:knowledge_base}:

\begin{equation} 
\label{eq:final}
\begin{split}
L_{final}(\bm{z}, \bm{t}, \bm{s}, \bm{c}; \bm{\theta}) = &L(\bm{z} + \bm{y}_{symb}, \bm{t}; \bm{\theta}) +
\\ 0.1 ~ &L_{sym}(\bm{s}, \bm{z} + \bm{y}_{symb}, \bm{t}; \bm{\theta}) + 0.1 ~ L_{obj}(\bm{c}, \bm{z} + \bm{y}_{symb}, \bm{t}; \bm{\theta})
\end{split}
\end{equation}

\section{Experimental Validation}
\label{sec:results}

We evaluate to what extent our proposed method is able to match an ad to its intended message (see Sec.~\ref{sec:task}).
We present the baselines against which we compare (Sec.~\ref{sec:baselines}), our metrics (Sec.~\ref{sec:metrics}), quantitative results on our main ranking task (Sec.~\ref{sec:quant}), results on QA as classification (Sec.~\ref{sec:qa_class}) and on three additional tasks (Sec.~\ref{sec:additional_tasks}).
Please see the supplementary file for implementation details, in-depth quantitative results, and qualitative results. 

\subsection{Baselines}
\label{sec:baselines}

We compare our \textsc{ADVISE} method (Sec.~\ref{sec:advise_final}) to the following approaches from recent literature. All methods are trained on the Ads Dataset \cite{Hussain_2017_CVPR}, using a train/val/test split of 60\%/20\%/20\%, resulting in around 39,000 images and more than 111,000 associated statements for training.

\begin{itemize}
\item \textsc{Hussain-Ranking} adapts \cite{Hussain_2017_CVPR}, the only prior method for decoding the message of ads. This method also uses symbol information, but in a less effective manner. The original method combines image, symbol, and question features, and trains for the 1000-way classification task. To adapt it,
we pointwise-add the image features (Inception-v4 as for our method) and symbol features (distribution over 53 predicted symbols), and embed them in 200-D using Eq.~\ref{eq:triplet} (using hard negative mining), setting $\bm{v}$ to the image-symbol feature. We tried four other ways (described in supp) of adapting \cite{Hussain_2017_CVPR} to ranking, but they performed worse. 

\item \textsc{VSE++} \cite{faghri2017vse++} (follow-up to \cite{kiros2014unifying}) uses the same method as Sec.~\ref{sec:triplet}. 
It is representative of one major group of recent image-text embeddings using triplet-like losses \cite{Niu_2017_ICCV,Mai_2017_CVPR,Karpathy_2015_CVPR,ren2016joint}.

\item \textsc{VSE}, which is like \textsc{VSE++} but without hard negative mining, for a more fair comparison to the next baseline. 

\item \textsc{2-way Nets} uses our implementation of \cite{Eisenschtat_2017_CVPR} (published code only demoed the network on MNIST) and is representative of a second  type of image-text embeddings using reconstruction losses \cite{Eisenschtat_2017_CVPR,Tsai_2017_ICCV}.

\end{itemize}

\subsection{Metrics}
\label{sec:metrics}

We compute two metrics: Rank, which is the averaged ranking value of the highest-ranked true matching statement (highest possible rank is 1, which means first place), and Recall@3, which denotes the number of correct statements ranked in the Top-3. We expect a good model to have low Rank and high Recall scores.
We use five random splits of the dataset into train/val/test sets, and show mean results and standard error over a total of 62,468 test cases (removing statements that do not follow the template ``I should [action] because [reason].'').

\subsection{Results on the main ranking task}
\label{sec:quant}

\begin{table*}[t]
    \centering
    \caption{Our main result. We show two methods that do not use hard negative mining, and three that do. Our method greatly outperforms three recent methods in retrieving matching statements for each ad. All methods are trained on the Ads Dataset of \cite{Hussain_2017_CVPR}. The best method is shown in \textbf{bold}, and the second-best in \emph{italics}}
    \begin{tabular}{|c||c|c||c|c|}
    \hline
    & \multicolumn{2}{|c||}{Rank (Lower $\downarrow$ is better)}     & \multicolumn{2}{|c|}{Recall@3 (Higher $\uparrow$ is better)}  \\
    \hline
    Method                      & PSA                           &   Product                     & PSA                               & Product               \\
    \hline                      
    \hline                      
    \textsc{2-way Nets}         & 4.836 ($\pm$ 0.090)           & 4.170 ($\pm$ 0.023)           & 0.923 ($\pm$ 0.016)               & 1.212 ($\pm$ 0.004)   \\
    \textsc{VSE}    & 4.155 ($\pm$ 0.091)           & 3.202 ($\pm$ 0.019)           & 1.146 ($\pm$ 0.017)               & 1.447 ($\pm$ 0.004)   \\
    \hline
    \textsc{VSE++}              & 4.139 ($\pm$ 0.094)           & 3.110 ($\pm$ 0.019)           & 1.197 ($\pm$ 0.017)               & 1.510 ($\pm$ 0.004)   \\
    \textsc{Hussain-Ranking}    & \emph{3.854} ($\pm$ 0.088)    & \emph{3.093} ($\pm$ 0.019)    & \emph{1.258} ($\pm$ 0.017)        & \emph{1.515} ($\pm$ 0.004)   \\
    \textsc{ADVISE} (ours)      & \textbf{3.013} ($\pm$ 0.075)  & \textbf{2.469} ($\pm$ 0.015)  & \textbf{1.509} ($\pm$ 0.017)      & \textbf{1.725} ($\pm$ 0.004) \\
    \hline
    \end{tabular} 
    \label{tab:main}
\end{table*}

We show the improvement that our method produces over state of the art methods, in Table \ref{tab:main}.
We show the better of the two alternative methods from Sec.~\ref{sec:knowledge_base}, namely \textsc{KB-Symbols}.
Since public service announcements (e.g. domestic violence or anti-bullying campaigns) typically use different strategies and sentiments than product ads (e.g. ads for cars or coffee), we separately show the result for PSAs and products. 
We observe that our method greatly outperforms the prior relevant research. PSAs in general appear harder than product ads (see Sec.~\ref{sec:intro}). 

\begin{figure}[t]
    \centering
    \includegraphics[width=1\linewidth]{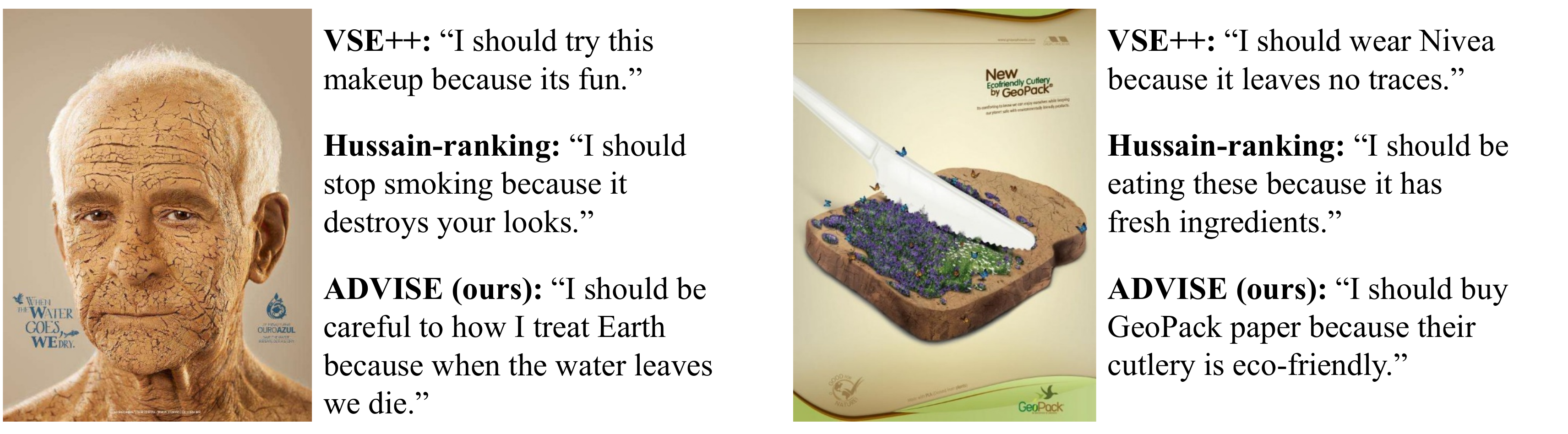}
    \caption{Our \textsc{ADVISE} method compared to the two stronger baselines. On the left, \textsc{VSE++} incorrectly guessed this is a makeup ad, likely because often faces appear in makeup ads. \textsc{Hussain-Ranking} correctly determined this is a PSA, but only our method was able to predict the topic, namely water/environment preservation. On the right, both \textsc{Hussain-Ranking} and our method recognized the concepts of freshness/naturalness, but our method picked a more specific statement. }
    \label{fig:qual_full}
\end{figure}

Compared to \textsc{2-way Nets} \cite{Eisenschtat_2017_CVPR}, \textsc{VSE} which does \emph{not} use hard negative mining is stronger by a large margin (14-23\% for rank, and 19-24\% for recall). \textsc{VSE++} produces more accurate results than both \textsc{2-way Nets} and \textsc{VSE}, but is outperformed by \textsc{Hussain-Ranking} and our \textsc{ADVISE}.
Our method is the strongest overall. It improves upon VSE++ \cite{faghri2017vse++} by 20-27\% for rank, and 14-26\% for recall. Compared to the strongest baseline, \textsc{Hussain-Ranking} \cite{Hussain_2017_CVPR}, our method is 20-21\% stronger in terms of rank, and 13-19\% stronger in recall.
Fig.~\ref{fig:qual_full} shows a qualitative result contrasting the best methods.

We also conduct ablation studies to verify the benefit of each component of our method. We show the \textsc{base triplet} embedding (Sec.~\ref{sec:triplet}) similar to \textsc{VSE++}; a \textsc{generic region} embedding using image regions learned using \cite{liu2016ssd} trained on the COCO \cite{lin2014microsoft} detection dataset;
\textsc{symbol region} embedding and \textsc{attention} (Sec.~\ref{sec:regions}); adding \textsc{symbol/object} constraints (Sec.~\ref{sec:symbols}); and including additive knowledge (Sec.~\ref{sec:knowledge_base}) using either \textsc{KB objects} or \textsc{KB symbols}.

\begin{table}[t]
    \centering
    \caption{(Left) Ablation study on PSAs. All external knowledge components except attention improve over basic triplet embedding. (Right) Ablation on products. General-purpose recognition approaches, e.g. regions and attention, produce the main boost}
    \begin{tabular}{|c|c|c|c|c||c|c|c|c|c|}
    \hline
                            & \multicolumn{4}{|c||}{PSA}                     & \multicolumn{4}{|c|}{Product}                 \\
                            \hline
                            &   &   & \multicolumn{2}{|c||}{\% improvement}  &   &   & \multicolumn{2}{|c|}{\% improvement}  \\
    Method                  & Rank $\downarrow$ & Rec@3 $\uparrow$  & Rank  & Rec@3 & Rank $\downarrow$ & Rec@3 $\uparrow$  & Rank  & Rec@3 \\     
    \hline
\textsc{base triplet}       & 4.139     & 1.197     &       &       & 3.110     & 1.510     &       &       \\
\textsc{generic region}     & 3.444     & 1.375     &   17  & 15    & 2.650     & 1.670     & 15    & 11    \\
\textsc{symbol region}      & 3.174     & 1.442     &   8   & 5     & 2.539     & 1.697     & 4     & 2     \\
+ \textsc{attention}        & 3.258     & 1.428     &   -3  & -1    & 2.488     & 1.726     & 2     & 2     \\
+ \textsc{symbol/object}    & 3.149     & 1.466     &   3   & 3     & 2.469     & 1.727     & 1     & $<$1    \\
+ \textsc{KB objects}       & 3.108     & 1.482     &   1   & 1     & 2.471     & 1.725     & $<$1  & $<$1    \\ 
+ \textsc{KB symbols}       & 3.013     & 1.509     &   4   & 3     & 2.469     & 1.725     & $<$1  & $<$1    \\
\hline
    \end{tabular}
    \label{tab:ablation}
\end{table}

The results are shown in Table \ref{tab:ablation} (left for PSAs, right for products). We also show percent improvement of each new component, computed with respect to the previous row, except for \textsc{KB objects} and \textsc{KB symbols}, whose improvement is computed with respect to the third-to-last row, i.e. the method on which both KB methods are based. The largest increase in performance comes from focusing on individual regions within the image. This makes sense because ads are carefully designed and multiple elements work together to convey the message. 
We see that these regions must be learned as visual anchors to symbolic concepts (\textsc{symbol region} vs \textsc{generic region}) to further increase performance. 

Beyond this, the story that the results tell differs between PSAs and products. Symbol/object constraints and additive branches are more helpful for the challenging, abstract PSAs that are the focus of our work. 
For PSAs, the additive inclusion of external information helps more when we directly predict the symbols (\textsc{KB symbols}), but also when we first extract objects and map these to symbols (\textsc{KB objects}). 
Note that \textsc{KB symbols} required 64,131 symbol labels. In contrast, \textsc{KB objects} relies on mappings between object and symbol words, which can be obtained more efficiently. While we obtain them as object-symbol similarities in our learned space, they could also be obtained from a purely textual, ad-specific resource. Thus, \textsc{KB objects} would generalize better to a new domain of ads (e.g. a different culture) where the data from \cite{Hussain_2017_CVPR} does not apply.

In Table \ref{tab:synonyms}, we show the object-symbol knowledge base that \textsc{KB objects} (Sec.~\ref{sec:knowledge_base}) uses. 
We show ``synonyms'' across three vocabularies: the 53 symbol words from \cite{Hussain_2017_CVPR}, the 27,999 words from the action/reason statements, and the 823 words from captions predicted for ads. We compute the nearest neighbors for each word in the learned space. This can be used as a ``dictionary'': If I see a given object, what should I predict the message of the ad is, or if I want to make a point, what objects should I use? 
\ak{In triplet ID 1, we see to allude to ``comfort,'' one might use a soft sofa.
From ID 2, if the statement contains ``driving,'' perhaps this is a safe driving ad, where visuals allude to safety and injury, and contain cars and windshields. 
We observe the different role of ``ketchup'' (ID 3) vs ``tomato'' (ID 4): the former symbolizes flavor, and the latter health. }

\begin{table}[t]
    \centering
    % \scriptsize
    \caption{Discovered synonyms between symbol, action/reason, and DenseCap words} 
    \begin{tabular}{|c|p{3.5cm}|p{4.4cm}|p{3.6cm}|}
    \hline
    ID & Symbol & Statement & DenseCap \\
    \hline
1 & \emph{comfort} & couch, sofa, soft & pillow, bed, blanket\\\hline
2 & safety, danger, injury & \emph{driving} & car, windshield, van\\\hline
3 & delicious, hot, food & \emph{ketchup} & beer, pepper, sauce\\\hline
4 & food, healthy, hunger & salads, food, salad & \emph{tomato} \\\hline
    \end{tabular}
    \label{tab:synonyms}
\end{table}

\begin{figure}[t]
    \centering
    \includegraphics[width=1.0\linewidth]{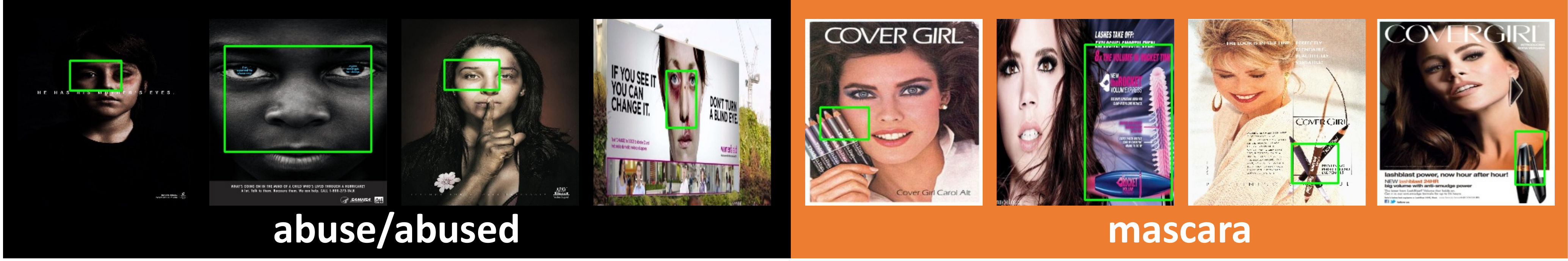}
    \caption{\ak{Application for ads image retrieval} (see  details in supp). \ak{We extract the CNN feature of each image region (Eq.~\ref{eq:region_embedding}), then use the word embeddings of ``abuse/abused'' and ``mascara'' to retrieve the most similar image regions (denoted using green boxes).}}
    \label{fig:qual_knn}
\end{figure}

In Fig.~\ref{fig:qual_knn}, we show the learned association between the individual words and symbolic regions. By learning from the ads image and statement pairs, our ADVISE model propagates words in the statement to the regions in the image thus associates each label-agnostic region proposal with  semantically meaningful words. At training time, we have neither box-level nor word-level annotations.

\subsection{Results on question-answering as classification }
\label{sec:qa_class}

For additional comparison to \cite{Hussain_2017_CVPR}, we evaluate our method on the question-answering task formulated as 1000-way single-word answer classification (Sec.~\ref{sec:task}). 
We now directly optimize for this classification task, but add our symbol-based region proposals, symbol/object constraints, and additive knowledge-based image representation. 
Our implementation of the method of Hussain et al. \cite{Hussain_2017_CVPR} pointwise-adds Inception-v4 image features and the symbol distribution, and obtains 10.03\% top-1 accuracy on PSAs, and 11.89\% accuracy on product ads (or 11.69\% average across ads regardless of type, which is dominated by product ads, and is close to the 11.96\% reported in \cite{Hussain_2017_CVPR}). 
Representing the image with a weighted summation of generic regions produced 10.42\% accuracy for PSAs, and 12.45\% for products (a 4\% and 5\% improvement, respectively). Using our method resulted in 10.94\% accuracy for PSAs, and 12.64\% for products (a 9\% and 6\% improvement over \cite{Hussain_2017_CVPR}, respectively). Note that a method known to work well for many recognition tasks, i.e. region proposals, leads to very small improvement in the case of QA classification for ads, so it is unlikely that any particular method would lead to a large improvement on this task. This is why we believe the ranking task we evaluate in Sec.~\ref{sec:quant} is more meaningful.

\begin{table}[t]
\centering
    \caption{Other tasks our learned image-text embedding helps with. We show rank for the first two (lower is better) and homogeneity \cite{rosenberg2007v} for the third (higher is better)}
   \begin{tabular}{|c|c|c|c|}
\hline
    Method                                      & Hard statements ($\downarrow$ better) & Slogans ($\downarrow$ better) & Clustering ($\uparrow$ better) \\     
    \hline
    \textsc{Hussain-Ranking}   & 5.595 ($\pm$ 0.027)           & 4.082 ($\pm$ 0.090)           & 0.291 ($\pm$ 0.002) \\
    \textsc{VSE++}                       & 5.635 ($\pm$ 0.027)           & 4.102 ($\pm$ 0.091)           & 0.292 ($\pm$ 0.002) \\
    \textsc{ADVISE} (ours)                      & \textbf{4.827} ($\pm$ 0.025)  & \textbf{3.331} ($\pm$ 0.077)  & \textbf{0.355} ($\pm$ 0.001) \\
    \hline
    \end{tabular}
    \label{tab:other}
\end{table}

\subsection{Results on additional tasks}
\label{sec:additional_tasks}

In Table \ref{tab:other}, we demonstrate the versatility of our learned embedding, compared to the stronger two baselines from Table \ref{tab:main}. None of the methods were retrained, i.e. we simply used the pre-trained embedding evaluated on statement ranking. First, we show a harder statement retrieval task: all statements that are to be ranked are from the same topic (e.g. all statements are about car safety or about beauty products). The second task uses creative captions that MTurk workers were asked to write for 2,000 ads in \cite{Hussain_2017_CVPR}. We rank these slogans, using an image as the query, and report the rank of the correct slogan. Finally, we check how well an embedding clusters ad images with respect to a ground-truth clustering defined by the topics of the ads. 

\section{Conclusion}
\label{sec:conclusion}

We presented a method for matching image advertisements to statements which describe the idea of the ad. Our method uses external knowledge in the form of symbols and predicted objects in two ways, as constraints for a joint image-text embedding space, and as an additive component for the image representation. We also verify the effect of state-of-the-art object recognition techniques in the form of region proposals and attention. Our method outperforms existing image-text embedding techniques \cite{Eisenschtat_2017_CVPR,faghri2017vse++} and a previous ad-understanding technique \cite{Hussain_2017_CVPR} by a large margin. Our region embedding relying on visual symbolic anchors greatly improves upon traditional embeddings. For PSAs, regularizing with external info provides further benefit.
In the future, we will investigate other external resources for decoding ads, such as predictions about the memorability or human attention over ads, and textual resources for additional mappings between physical and abstract content. We will use our object-symbol mappings to analyze the visual variability  the same object category exhibits when used for different ad topics.

\section*{Acknowledgments}
\label{sec:ack}

This material is based upon work supported by the National Science Foundation under Grant Number 1566270. This research was also supported by an NVIDIA hardware grant. Any opinions, findings, and conclusions or recommendations expressed in this material are those of the author(s) and do not necessarily reflect the views of the National Science Foundation. We thank the anonymous reviewers for their feedback and encouragement. 
%
% ---- Bibliography ----
%
% BibTeX users should specify bibliography style 'splncs04'.
% References will then be sorted and formatted in the correct style.

\bibliographystyle{splncs04}
\bibliography{refs_main,refs_workshop}

\end{document}

\pagestyle{headings}

\title{ADVISE: Symbolism and External Knowledge for Decoding Advertisements\\(Supplementary Material)}

\titlerunning{ADVISE: Symbolism and External Knowledge for Decoding Advertisements}

%\author{Keren Ye\orcidID{0000-0002-7349-7762} \and
%Adriana Kovashka\orcidID{0000-0003-1901-9660}}
\author{Keren Ye \and Adriana Kovashka}

\authorrunning{Keren Ye \and Adriana Kovashka}

\institute{University of Pittsburgh, Pittsburgh PA 15260, USA\\
\email{\{yekeren,kovashka\}@cs.pitt.edu}}

\maketitle

% ----- INTRO

In this document, we include more information and statistics about our ADVISE model and some implementation decisions. We also provide additional quantitative and qualitative experimental results. 

We first describe in more depth the implementation and evaluation setup. We provide details about the implementation and training of our ADVISE model in Sec.~\ref{sec:advise_details}. 
In Sec.~\ref{sec:hussain}, we describe different ways of adapting Hussain et al. [22]'s method to the ranking task. 
In Sec.~\ref{sec:reason_50_statements}, we explain the reason why we choose to rank 50 statements for our main task. 

Next, we provide additional quantitative results which demonstrate the contribution of different algorithmic choices that we made.
In Sec.~\ref{sec:different_samplings}, we provide details justifying our choice of the value of $k'$ for hard negative mining. 
In Sec.~\ref{sec:different_attn}, we demonstrate different strategies and justify our choice of the attention mechanism used.
In Sec.~\ref{sec:different_proxies}, we quantitatively demonstrate that it is not sufficient to use any type of label in the domain of interest for the method component described in Sec. 3.4 of the main text; in particular, we show that using the symbol labels as constraints gives more improvement than using topic labels. 
In Sec.~\ref{sec:in_depth_quant}, we break down the ranking task evaluation into topics. 

Finally, to enable a more intuitive understanding of our model, we provide more qualitative results of the ranking task in Sec.~\ref{sec:ranking_quali}, including both statement ranking and hard-statement ranking. In Sec.~\ref{sec:region_knn}, we show qualitatively that the ADVISE model learns not only the image representation but also meaningful \emph{region} representations.

\clearpage

% ----- DATA

% IMPLEMENTATION DETAILS OF THE ADVISE MODEL
\section{Training the ADVISE model}
\label{sec:advise_details}

As shown in Figure 2 in our paper, there are two branches in our model, the main branch (top left), and the knowledge embedding branch (top right). We do not train the branches jointly at the very beginning, instead, we at first train the main branch till the model converges, then we add the knowledge base information. It is beneficial to incorporate knowledge additively using this two-step process. In our experiments, training the knowledge branch requires less time compared to training the main branch. Thus, one could efficiently experiment with multiple extra types of knowledge given that the main branch is trained and the entry-points of symbols are properly set. Another advantage of the two-step training is that the knowledge branch serves a role similar to a residual branch, that is it would not hurt the performance of the existing main branch.

We experimented with using Adagrad, Adam, RMSProp and found Adagrad to give the best results. We use a learning rate of 2.0 and apply no decay strategy on the learning rate. Also, we did not use different gradient multipliers for image and text embedding networks. For both the image embedding network ($\bm {w}$ in Eq. 4) and the attention prediction network ($\bm{w_a}$ in Eq. 5), we use batch normalization layer, a weight decay of 1e-6, and dropout keep probability of 0.7 for the input Inception V4 feature. For the text embedding network of the statement (Sec. 3.2), DenseCap captions, and symbols (Sec. 3.4), we use a weight decay of 1e-8 and a dropout keep probability of 0.7 for the embedding weights. The DenseCap captions do not share weights with the statements, but they are both initialized from GLOVE word embedding [48]. For the ``unknown'' words of DenseCap captions, Ads statements and the symbol embedding vectors, they are initialized from the uniform distribution ranging from -0.08 to 0.08. According to our experiments, adding a dropout layer (with keep probability of 0.5) after the pointwise multiplication of image and text embedding of $||x-y||_2^2$ is really important. It assures that the model will not overfit on the training set. For the triplet training, we mine the most challenging 32 negative examples in the 128-sized training batch, we weigh 0.1 for the symbol loss and object loss as mentioned in Eq. 8. Based on the settings mentioned above, we train the main branch for 100,000 steps and use Recall@3 as the metric to choose the best model on the validation set.

We build the knowledge branch after getting the checkpoint of the main branch. During this second phase, we freeze all of the parameters we harvest in the first step, saying $\bm{w}$ for the image embedding $\bm{z}$, $\bm{w_a}$ for attention prediction, $\bm{t}$ for Ads statement embedding, $\bm{c}$ for DenseCap embedding, and $\bm{s}$ for symbol embedding. In the meantime, we also freeze the parameters of the pre-trained symbol classifiers ($\bm{u_{symb}}$ as mentioned in Sec. 3.5) since the classifiers are part of our prior knowledge. Therefore, the only parameters of our ADVISE model in the second training phase are the 53 scalar values, that is, $\alpha_j^{symb}$ (for each of $j \in \{1,...,C=53\}$), denoting the importance of the 53 classifiers. If the main branch captures all the information during the training, assigning 0 to all $\alpha_j^{symb}$ would not hurt the performance of the model. In case the main branch does \emph{not} capture all relevant information, the knowledge branch may provide complementary information to help to improve the final performance. Thus the knowledge branch trained is similar to a residual branch. Please note that the symbol embedding $\textbf{\emph{s}}$ is learned in the first phase using symbol constraint, and so these symbol embedding vectors serve as entry points for external knowledge. In order to bound the $\alpha_j^{symb}$, we apply the $sigmoid$ activation on them and multiply them by 2. Thus the 53 confidence scores of the classifiers are ranging from 0 to 2, in which 0 means the associated classifier is not useful. To train the knowledge branch, we use the Adam optimizer, and a learning rate of 0.01. Based on this setting, we train both the branches jointly (with main branch freeze) for 5,000 steps and cross-validate to get the best model.

\begin{figure}[h]
    \centering
    \includegraphics[width=0.6\linewidth]{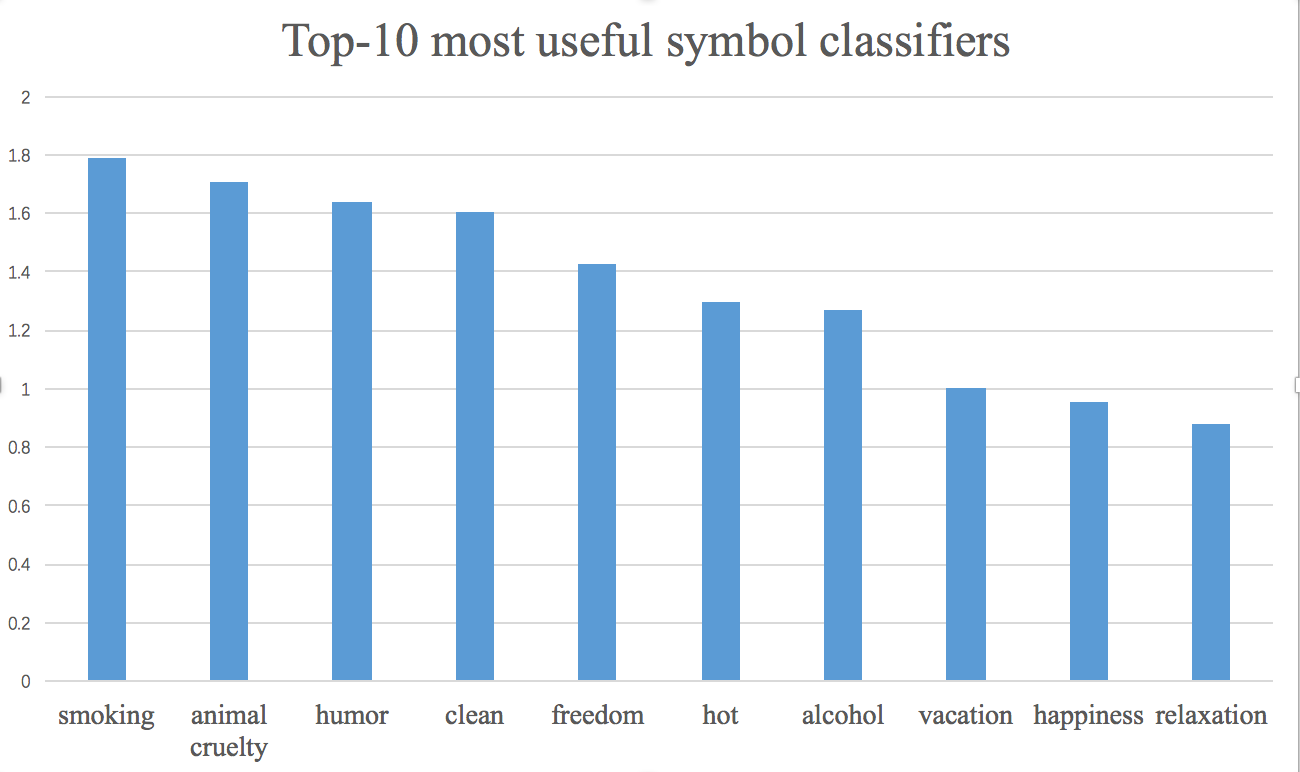}
    \includegraphics[width=0.6\linewidth]{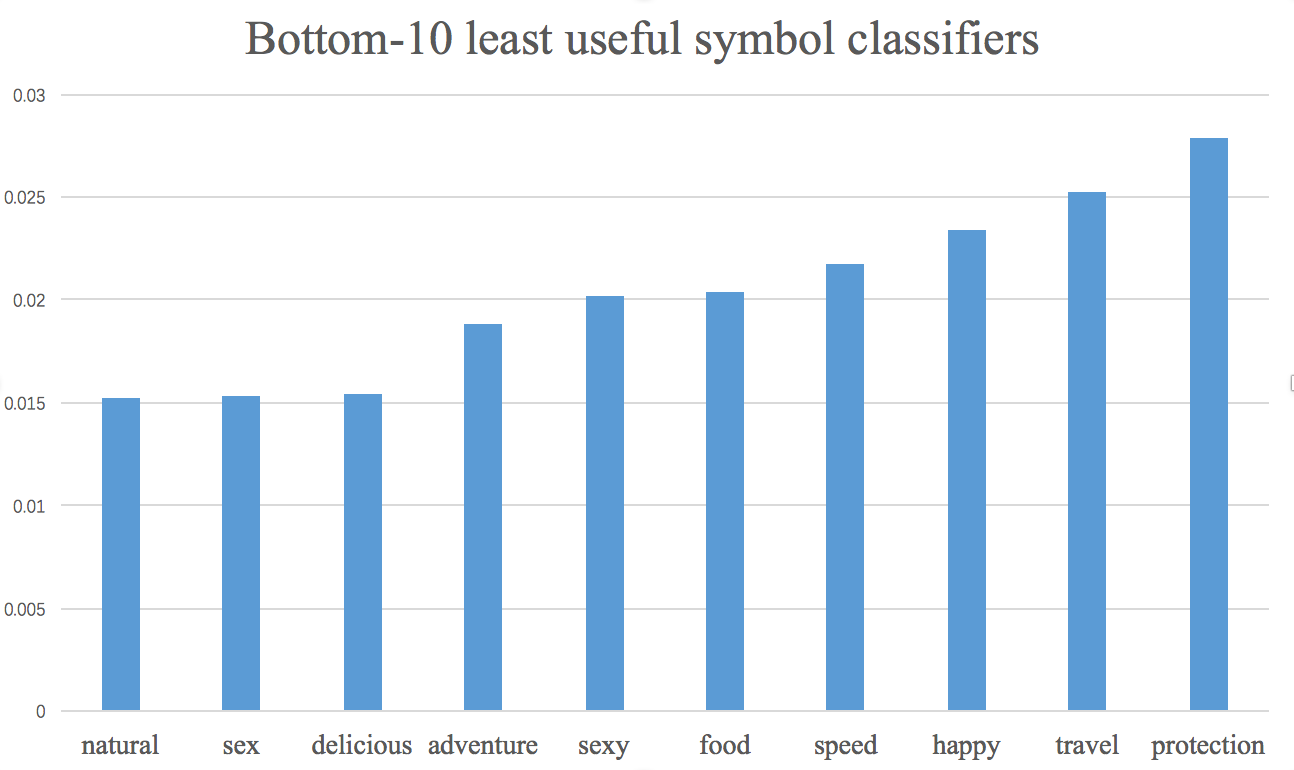}
    \caption{Confidence scores of the symbol classifiers.}
    \label{fig:symbol_classifiers}
\end{figure}

In order to show that our ADVISE really learns the importance of different classifiers, we show the confidence scores of the top-10 most useful and bottom-10 least useful symbol classifiers in Figure~\ref{fig:symbol_classifiers}. The figure accords to our expectation because our ADVISE model tends to weigh more on classifiers such as ``smoking'', ``animal cruelty'' while in the Ads dataset there are only a limited number of training examples for these symbols. Thus incorporating knowledge for these not well-trained symbols is necessary.

\section{Adapting Hussain et al.'s method [22]}
\label{sec:hussain}

Hussain et al. [22] developed the only method we are aware of for understanding the messages of advertisements.
In the main text, we show the results of the most promising way of adapting Hussain et al.'s question-answering method for our ranking task. 
The other ways of adapting Hussain's method that we tried include retrieving the statement that had the highest similarity between the single-word picked in [22] and the action-reason statements, using a standard GLOVE embedding or the embedding learned in our method. We also tried concatenating the Inception-V4 and symbol features rather than pointwise-adding them, and using the original VGG features used in [22].

% EXPLANATION FOR RANKING of 50 STATEMENTS
\section{Reason to choose 50 statements}
\label{sec:reason_50_statements}

Our evaluation is similar to [3] which provides 18 choices.
The task is challenging, and enlarging the list would make all methods score so poorly (e.g. low Recall@3) that their performance would be hard to compare. 
Table ~\ref{tab:stmt-cluster} demonstrates this challenge and shows that the 3 correct statements are internally more similar than a correct and an incorrect statement.
We computed $D_{within}^{min}$, $D_{between}^{min}$, $D_{within}^{avg}$, $D_{between}^{avg}$
as follows, where
$\bm{x}$ is an image in Ads dataset $\mathbb{D}$, $P(\bm{x})$ is the set of related statements for $\bm{x}$,  $N(\bm{x})$ are the randomly sampled statements from images other than $\bm{x}$, and $g(\cdot)$ computes average GLOVE embedding.

\begin{equation*} \label{eq:distance}
\begin{split}
D_{within}^{min}&=\underset{x\in \mathbb{D}}{\avg}\min\limits_{a,b\in P(x), a\ne b} ||g(a)-g(b)||_2^2 \\
D_{within}^{avg}&=\underset{x\in \mathbb{D}}{\avg}\underset{~a,b\in P(x), a\ne b}{\avg} ||g(a)-g(b)||_2^2 \\
D_{between}^{min}&=\underset{x\in \mathbb{D}}{\avg}\min\limits_{a\in P(x), b\in N(x)} ||g(a)-g(b)||_2^2 \\
D_{between}^{avg}&=\underset{x\in \mathbb{D}}{\avg}\underset{~a\in P(x), b\in N(x)}{\avg} ||g(a)-g(b)||_2^2 \\
\end{split}
\end{equation*}

\begin{table}[h]
    \centering
    \caption{The reason to choose 50 statements. We compute L2 distance between statements belonging to same (within) and different images (between). With more candidates, sampled negatives are hard to distinguish: the min ``between'' distance becomes similar to the ``within'' distance.} %($D_{between}^{min}$ becomes smaller).}
    %difference in the minimum distance between related statements and randomly sampled statements become small.}
    \begin{tabular}{|c|c|c|c|c|}
    \hline
    \# statements & $D_{within}^{min}$ & $D_{within}^{avg}$ & $D_{between}^{min}$ & $D_{between}^{avg}$ \\
    \hline
    \textsc{10}  & \multirow{ 3}{*}{1.395} & \multirow{ 3}{*}{1.551} & 1.619 & 1.937\\
    %\textsc{20}  &  &  & 1.526 & 1.938 \\
    \textsc{50}  &  &  & 1.437 & 1.938 \\
    %\textsc{100} &  &  & 1.379 & 1.937 \\
    \textsc{200} &  &  & 1.324 & 1.938 \\
    \hline
    \end{tabular}
    \label{tab:stmt-cluster}
\end{table}

% COMPARE SAMPLING STRATEGIES

\section{Hard negative mining}
\label{sec:different_samplings}

% DIFFERENT TYPES OF PROXIES

\ak{In Table \ref{tab:neg-mining}, we show that negative mining strategy does matter in our task. Using 32 hard negatives is better than using all or just the most challenging example. }

\begin{table*}[h!]
    \centering
    \caption{Hard negative mining with different top-k hyperparam, using batch size 128.}
    \begin{tabular}{|c||c|c||c|c|}
    \hline
    & \multicolumn{2}{|c||}{Rank (Lower $\downarrow$ is better)}     & \multicolumn{2}{|c|}{Recall@3 (Higher $\uparrow$ is better)}  \\
    \hline
    Method & PSA & Product & PSA & Product \\
    \hline                      
    \hline                      
    \textsc{1 negative} & 6.033 ($\pm$ 0.127) & 4.647 ($\pm$ 0.026) & 0.853 ($\pm$ 0.015) & 1.091 ($\pm$ 0.003)\\
    \textsc{32 negatives (VSE++)} & \textbf{4.139} ($\pm$ 0.094) & \textbf{3.110} ($\pm$ 0.019) & \textbf{1.197} ($\pm$ 0.017) & \textbf{1.510} ($\pm$ 0.004)   \\
    \textsc{All negatives (VSE)} & 4.155 ($\pm$ 0.091) & 3.202 ($\pm$ 0.019) & 1.146 ($\pm$ 0.017) & 1.447 ($\pm$ 0.004)  \\
    \hline
    \end{tabular}
    
    \label{tab:neg-mining}
\end{table*}

% COMPARE DIFFERENT TYPES OF PROXIES
\section{Region-based v.s. standard attention}
\label{sec:different_attn}

We wish to verify that bottom-up region-based attention is more appropriate for our task than standard attention.
\ak{We refer to the AMC model [6] to implement the baseline standard attention mechanism: we divide the image into $3\times3$ grids, apply Inception-v4 [58] to get features per cell, obtain regional features (including original image as proposal), then predict attention distribution. Thus we keep the basic feature, number of regions, and resolution the same as for our method. Table ~\ref{tab:attention} shows two ablations of our method and standard attention, for three tasks. 
Note \textsc{+attention} from the main text is not our contribution; \textsc{symbol region} is.
Standard attention is inferior. For PSAs, which is our focus, our symbol-based attention is the strongest attention method overall.}

\begin{table*}[h!]
    \centering
    \caption{Region-based vs. standard image attention. \textsc{symbol region} uses mean pooling, \textsc{region-based att} (\textsc{+attention} in Tab.2 of paper) uses attention pooling over region features, \textsc{standard att} applies attention pooling on evenly split $3\times3$ grids.}
    \begin{tabular}{|c|||c|c||c|c|||c|c||c|c|||c|}
    \hline
    & \multicolumn{4}{|c|||}{Statement} & \multicolumn{4}{c|||}{Slogan} & Clustering\\
    \hline
    & \multicolumn{2}{c||}{Rank $\downarrow$  }     & \multicolumn{2}{|c|||}{Recall@3  $\uparrow$ } & \multicolumn{2}{c||}{Rank $\downarrow$ }     & \multicolumn{2}{|c|||}{Recall@3 $\uparrow$ }  & Homogen $\uparrow$ \\
    \hline
    Method & PSA & Product & PSA & Product & PSA & Product & PSA & Product & All \\
    \hline                      
    \hline          
    \textsc{symbol region}      & \textbf{3.174} & 2.539 & \textbf{1.442} & 1.697   & \textbf{3.774} & 3.344 & 1.121 & 1.182 & 0.331 \\   
    \textsc{region-based att}   & 3.258 & 2.488 & 1.428 & \textbf{1.726}            & 3.850 & \textbf{3.257} & \textbf{1.155} & \textbf{1.205} & \textbf{0.355} \\
    \hline
    \textsc{standard att}       & 3.382 & \textbf{2.482} & 1.415 & 1.720            & 3.954 & 3.320 & 1.073 & 1.185 & 0.339 \\    
    
    \hline
    \end{tabular}
    \label{tab:attention}
\end{table*}

% COMPARE DIFFERENT TYPES OF PROXIES
\clearpage
\section{Different types of proxies}
\label{sec:different_proxies}

% DIFFERENT TYPES OF PROXIES

In Table \ref{tab:type_label} and Table \ref{tab:type_label_improve}, we show that not any type of label would suffice as constraint. In particular, the Ads Dataset includes 6 times more topic labels that could be used as constraints compared to symbol labels (64,325\footnote{The Ads dataset [22] involves 64,832 images with topic annotations. However, the actual topic labels (64,325) we can use as the constraint is the intersection of the topic annotations and the format-filtered (Sec. 4.2) statement annotations.} vs 10,493\footnote{There are 13,938 images annotated with symbols in Ads dataset [22], and we use these images to train our region proposal network in Sec. 3.3. However, not all of the free-formed symbol annotations could be transformed into the 53 symbols that [22] used. 10,493 is the number of images that have symbol annotations for the 53 symbols.}
images annotated; almost every image is annotated with topics yet only around 15\% images are annotated with symbols). Despite this, 
symbol labels give much greater benefit. Thus, [44]'s proxy approach is not enough; the type of labels must be carefully chosen. 

\label{sec:quant}

\begin{table*}[h]
    \centering
    \caption{Different types of labels as constraints. The baseline method (``No extra components'') uses image attention (Sec. 3.3) but does not have the components from Sec. 3.4-3.5 of the main text.}
    \begin{tabular}{|c|c|c||c|c|}
    \hline
    & \multicolumn{2}{|c||}{Rank (Lower $\downarrow$ is better)}     & \multicolumn{2}{|c|}{Recall@3 (Higher $\uparrow$ is better)}  \\
    \hline
    Method                      & PSA                           &   Product                     & PSA                               & Product               \\
    \hline                      
    \hline
No extra components & 3.258 ($\pm$ 0.081) & 2.488 ($\pm$ 0.015) & 1.428 ($\pm$ 0.017) & 1.726 ($\pm$ 0.004) \\
Symbol labels & 3.171 ($\pm$ 0.081) & 2.465 ($\pm$ 0.015) & 1.471 ($\pm$ 0.017) & 1.726 ($\pm$ 0.004) \\
Topic labels & 3.186 ($\pm$ 0.079) & 2.477 ($\pm$ 0.015) & 1.456 ($\pm$ 0.017) & 1.728 ($\pm$ 0.004) \\
    \hline
    \end{tabular}
    \label{tab:type_label}
\end{table*}

\begin{table}[]
    \centering
    \caption{\% improvement for different types of labels as constraints.} 
    \begin{tabular}{|c|c|c|c|c|}
    \hline
    & \multicolumn{2}{|c|}{PSA} & \multicolumn{2}{|c|}{Product}\\
    \hline
    Method                      & Rank $\downarrow$ & Rec@3 $\uparrow$  & Rank $\downarrow$     & Rec@3 $\uparrow$ \\     
    \hline
    Symbol labels               & 3                 & 3                 & 1                     & 0           \\
    Topic labels                & 2                 & 2                 & 0                     & 0           \\
    \hline
    \end{tabular}
    \label{tab:type_label_improve}
\end{table}

\clearpage
\section{In-depth quantitative results}
\label{sec:in_depth_quant}

\noindent \textbf{Quantitative results with extra measurement.}
We provide Rank and Recal@3 as the measurement of our model in the paper. In Table \ref{tab:in_depth_quant}, we also compute three other metrics: Recall@10, which denotes the number of correct statements ranked in the Top-10; RankAvg, the average ranking value of the averaged-ranked true matching statement; RankMedian, the average ranking value of the median-ranked true matching statement.

The reason that we only use Rank in the paper instead of RankAvg or RankMedian is that we found that in the Ads dataset [22], there are some really noisy annotations that ruin the metrics of RankAvg and RankMedian. This could be seen from Figure~\ref{tab:in_depth_quant} where RankAvg is always worse than RankMedian.
\begin{table*}[h]
    \centering
    \caption{Our main result with extra measurements. The best method is shown in \textbf{bold}. For Recall@3 and Recall@10, higher values ($\uparrow$) are better. For Rank, RankAvg, and RankMedian, lower values ($\downarrow$) are better.} 
    \begin{tabular}{|c|l||c|c|c|}
    \hline
     & & \textsc{VSE++} & \textsc{Hussain-Ranking} & \textsc{ADVISE} (ours) \\
    \hline
    \hline
 \multirow{ 4}{*}{Product}  & $\uparrow$ Recall@3 & 1.510 ($\pm$ 0.004) & 1.515 ($\pm$ 0.004) & \textbf{1.725} ($\pm$ 0.004)\\
 & $\uparrow$ Recall@10 & 2.379 ($\pm$ 0.003) & 2.392 ($\pm$ 0.003) & \textbf{2.527} ($\pm$ 0.003)\\
 \cline{2-5}
 & $\downarrow$ Rank & 3.110 ($\pm$ 0.019) & 3.093 ($\pm$ 0.019) & \textbf{2.469} ($\pm$ 0.015)\\
 & $\downarrow$ RankAvg & 7.311 ($\pm$ 0.029) & 7.122 ($\pm$ 0.028) & \textbf{6.143} ($\pm$ 0.025)\\
 & $\downarrow$ RankMedian & 6.392 ($\pm$ 0.030) & 6.297 ($\pm$ 0.029) & \textbf{5.252} ($\pm$ 0.026)\\
 \hline
 \hline
 \multirow{ 4}{*}{PSA} & $\uparrow$ Recall@3 & 1.197 ($\pm$ 0.017) & 1.258 ($\pm$ 0.017) & \textbf{1.509} ($\pm$ 0.017)\\
 & $\uparrow$ Recall@10 & 2.089 ($\pm$ 0.017) & 2.151 ($\pm$ 0.017) & \textbf{2.323} ($\pm$ 0.015)\\
 \cline{2-5}
 & $\downarrow$ Rank & 4.139 ($\pm$ 0.094) & 3.854 ($\pm$ 0.088) & \textbf{3.013} ($\pm$ 0.075)\\
 & $\downarrow$ RankAvg & 9.424 ($\pm$ 0.135) & 8.718 ($\pm$ 0.127) & \textbf{7.553} ($\pm$ 0.119)\\
 & $\downarrow$ RankMedian & 8.268 ($\pm$ 0.143) & 7.741 ($\pm$ 0.135) & \textbf{6.394} ($\pm$ 0.121)\\
    \hline
    \end{tabular}
    \label{tab:in_depth_quant}
\end{table*}

\begin{figure}[t]
    \centering
    \includegraphics[width=1\linewidth]{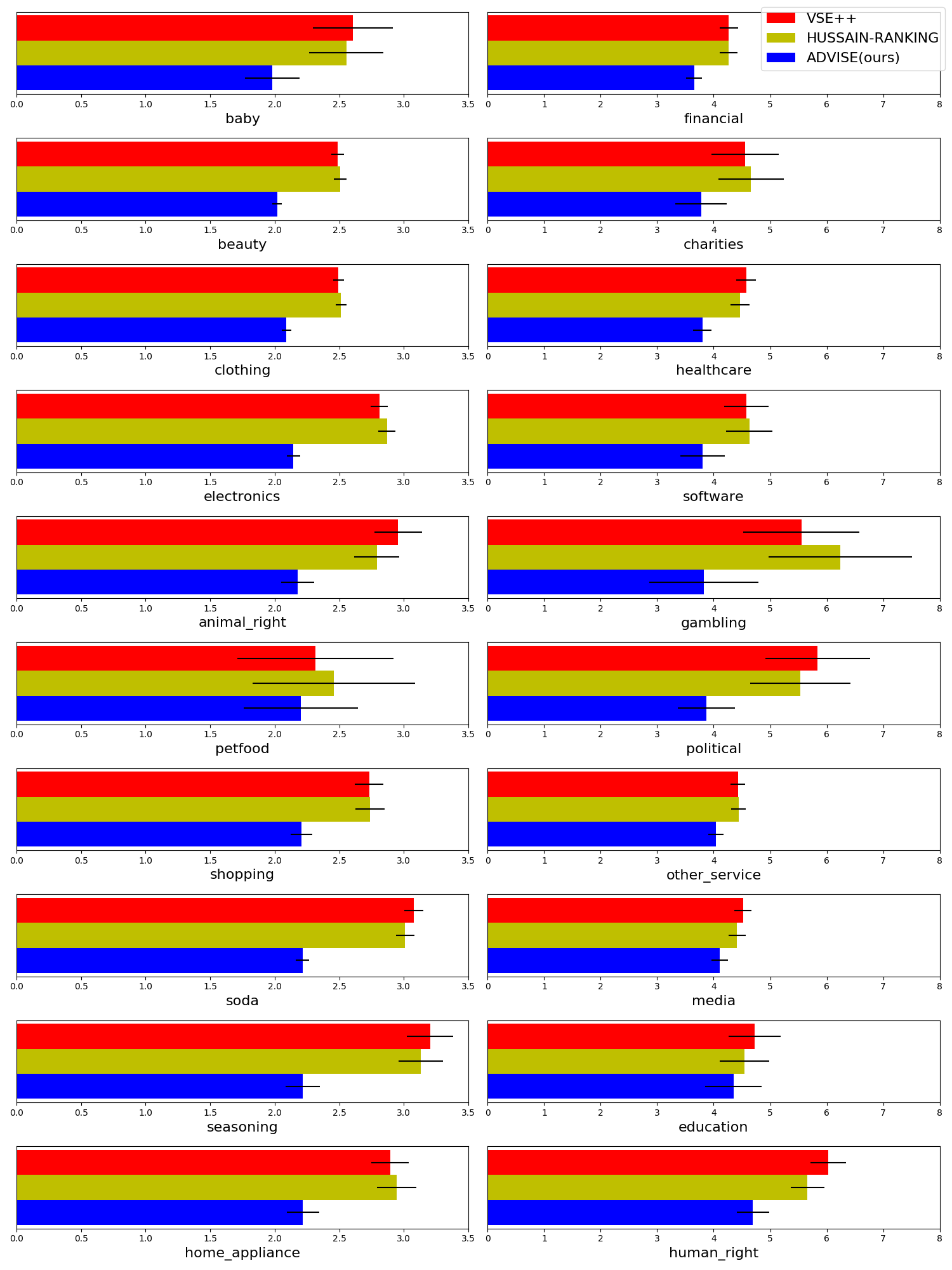}
    \caption{Bar chart of the statement ranking task. The x-axis denotes the Rank measurement. Lower is better.  The error bar shows the standard error, which is defined as $\delta/\sqrt{n}$ where $\delta$ is the standard deviation of the Rank and $n$ is the number of examples. Our ADVISE model is always better than the other two baselines.}
    \label{fig:per_topic_quant}
\end{figure}

\begin{figure}[t]
    \centering
    \includegraphics[width=1\linewidth]{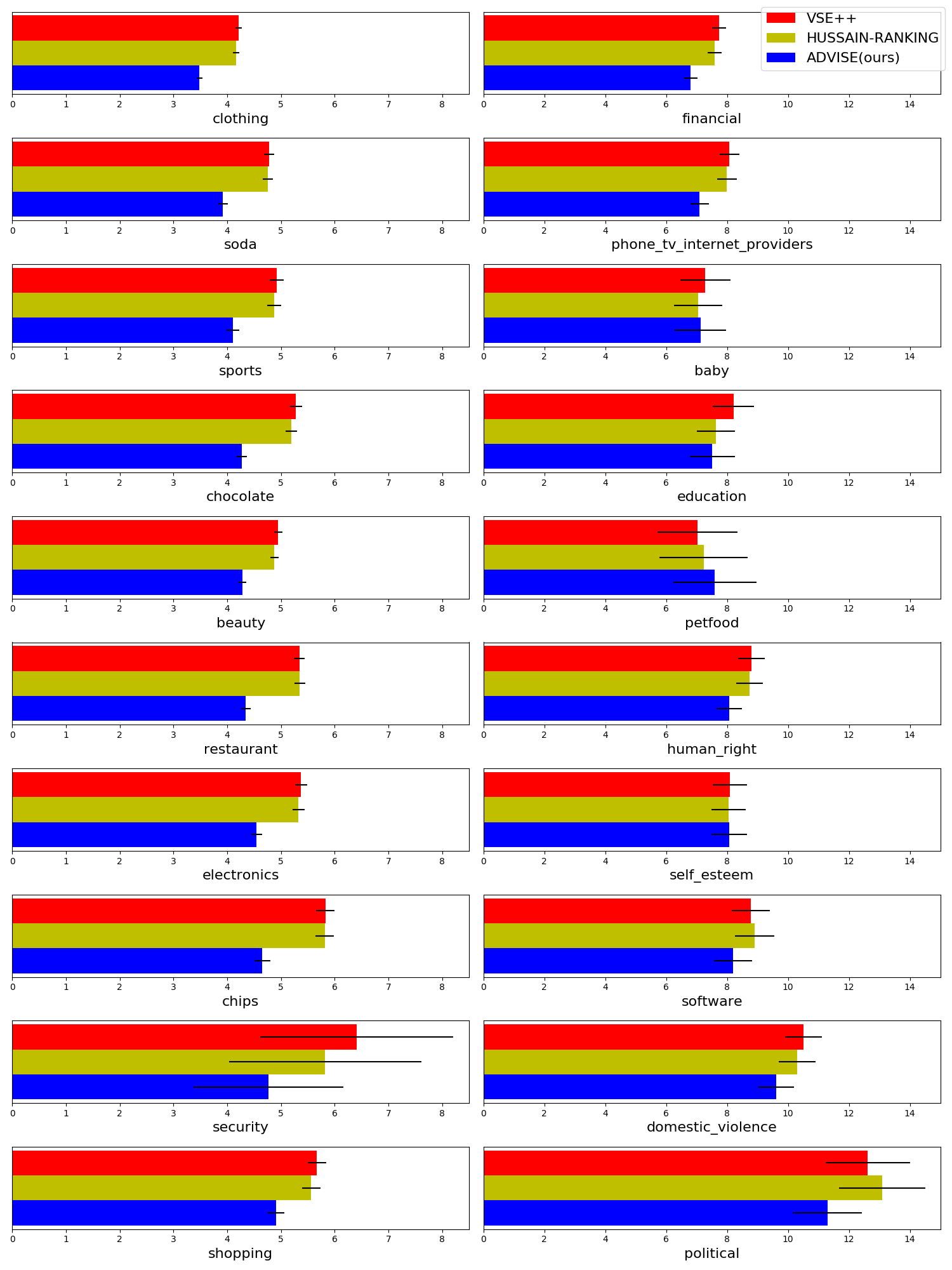}
    \caption{Bar chart of the hard-statement ranking task. The x-axis denotes the Rank measurement. The error bar shows the standard error, which is defined as $\delta/\sqrt{n}$ where $\delta$ is the standard deviation of the Rank and $n$ is the number of examples. Our ADVISE model is better than the two baselines in most of the cases.}
    \label{fig:per_topic_quant_hard}
\end{figure}

\noindent \textbf{Per-topic evaluation.}
We provide per-topic quantitative results to further compare our ADVISE model and the two strong baselines: \textsc{VSE++} [11] and \textsc{Hussain-Ranking} [22]. Figures~\ref{fig:per_topic_quant} and \ref{fig:per_topic_quant_hard} show the per-topic evaluation results of the statement ranking and hard-statement ranking task respectively. In both figures, we show bar charts of the best performing 10 topics (left) and the worst performing 10 topics (right), in terms of the Rank measurement of our ADVISE model (shown in blue).

Intuitively, the hard-statement ranking task defined in our paper emphasizes more the ability to distinguish the within-topic nuances, yet our main ranking task focus on both the within-topic and between-topic differences. We see from Figures~\ref{fig:per_topic_quant} and \ref{fig:per_topic_quant_hard} that ads of different topics perform differently on the two tasks. For example ``baby'' (baby products) ranks the first in the main ranking task, yet it is also in the worst-10 in terms of Rank in the hard-statement ranking task. Our explanation is that ``baby'' ads are quite distinguishable from others, yet they do not have sub-categories within themselves. Another example is ``clothing'', which has good performance on both tasks. The reason is that ``clothing'' ads are distinguishable from others, while they can still be further classified as, e.g., ``jean'', ``watches'', ``outfit'', and so on.

\clearpage
\section{Ranking task - qualitative results}
\label{sec:ranking_quali}

\begin{figure}[t]
    \centering
    \includegraphics[width=1\linewidth]{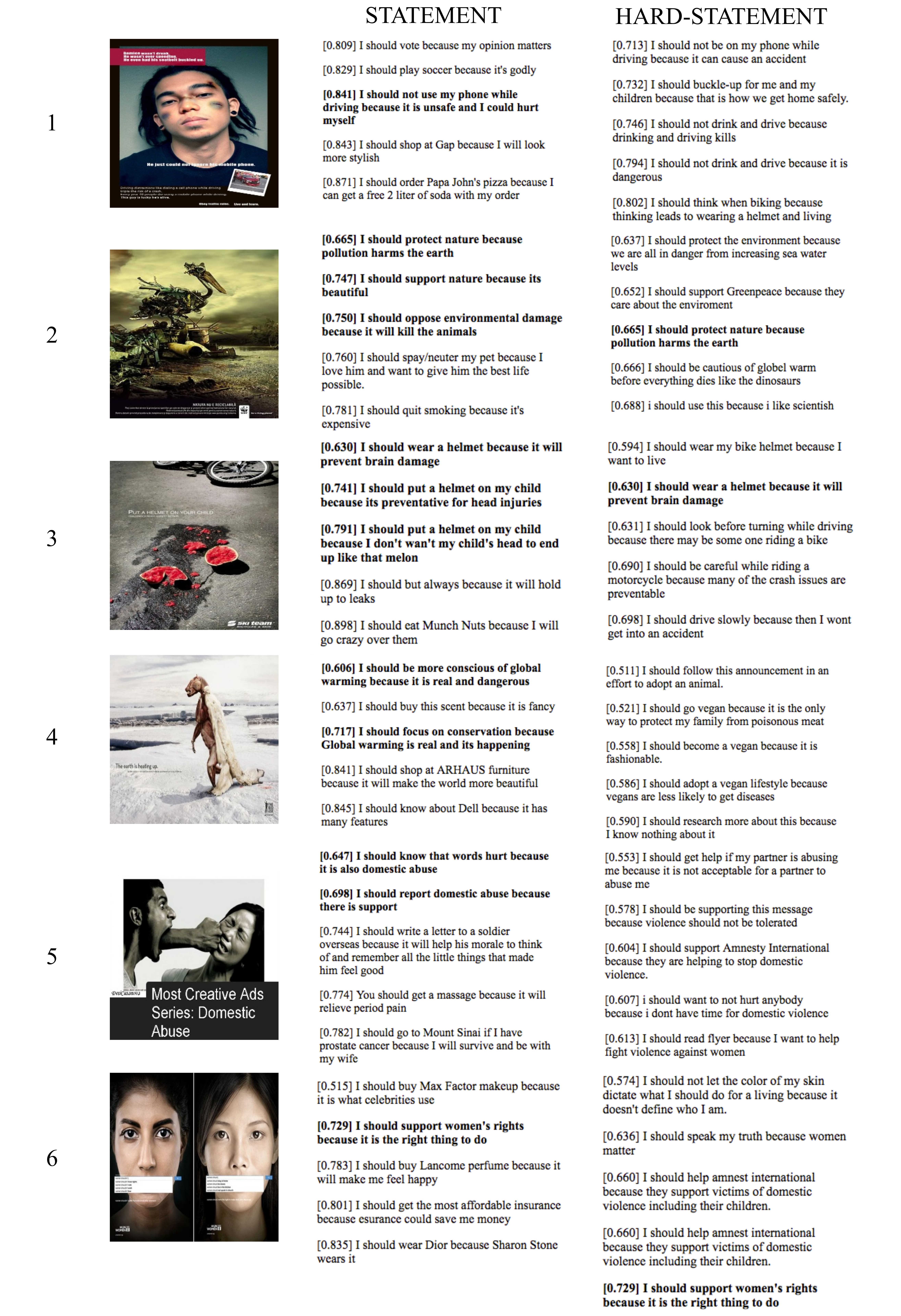}
    \caption{Statement/Hard-statement ranking task of PSA ads. We show the top-5 multiple-choice answers ranked by our ADVISE model for both statement and hard-statement ranking task. Statements in \textbf{bold} are the correct predictions.}
    \label{fig:psa_ranking}
\end{figure}

\begin{figure}[t]
    \centering
    \includegraphics[width=1\linewidth]{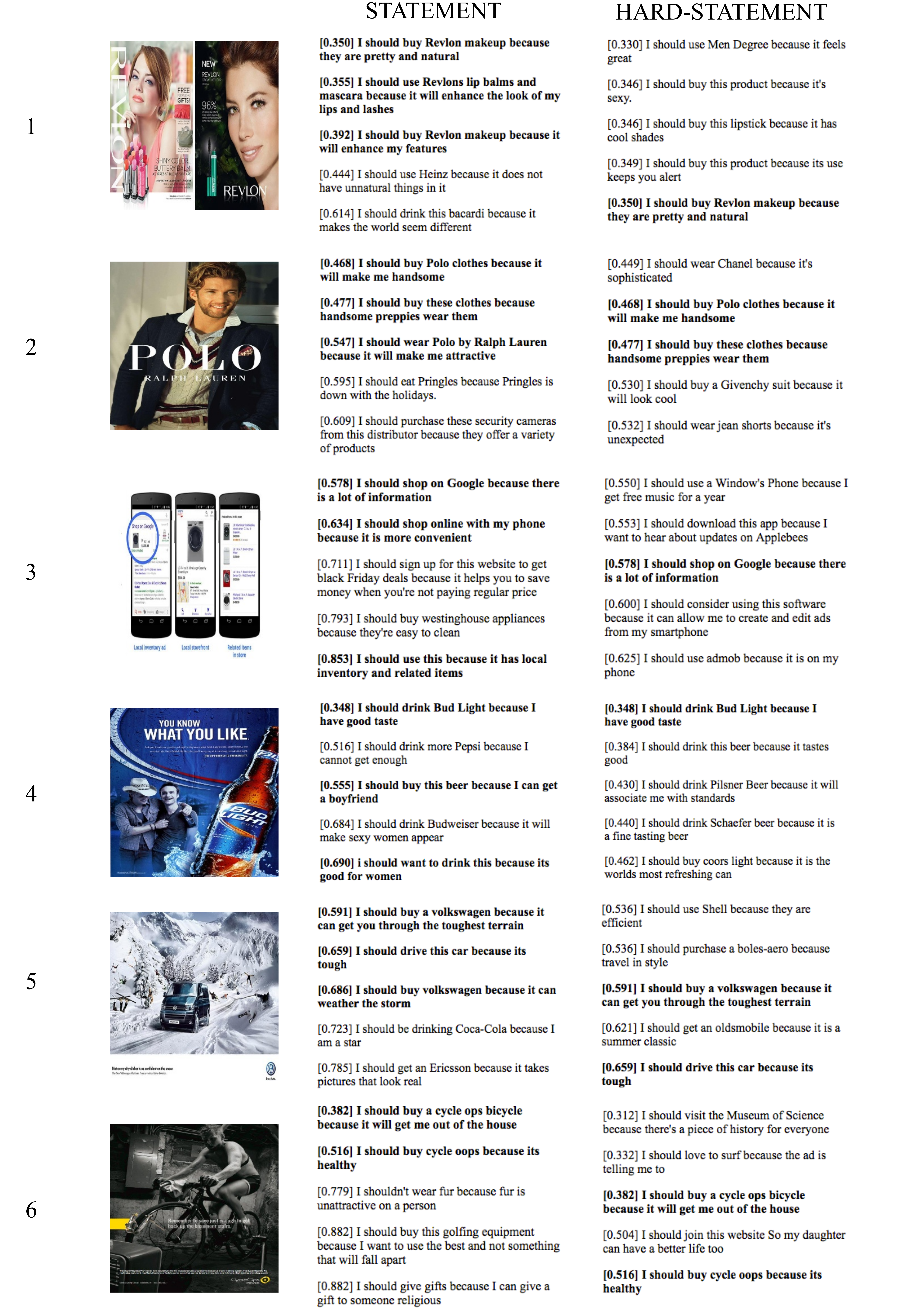}
    \caption{Statement/Hard-statement ranking task of product ads. We show the top-5 multiple-choice answers ranked by our ADVISE model for both statement and hard-statement ranking task. Statements in \textbf{bold} are the correct predictions.}
    \label{fig:product_ranking}
\end{figure}

We provide more qualitative examples of both the statement ranking task and the hard-statement ranking task for both PSAs (Figure \ref{fig:psa_ranking}) and product ads (Figure ~\ref{fig:product_ranking}). We can see from both figures, that hard-statement ranking task requires the model to have a deep understanding not only about the topic and the purpose of the ads but also the details such as the brand of product, the reasoning of causal relation, etc. We also see from the qualitative examples that failure in the hard-statement ranking task does not always mean we failed to understand the ads. For example, the top-1 hard-statement prediction of row 1 of Figure~\ref{fig:psa_ranking} has already captured all of the information in the ad, ``I should not be on my phone while driving because it can cause an accident'', yet this statement is from another similar image which causes the evaluation to not count this prediction as correct. %Therefore, we believe the (simpler) statement ranking task which does not restrict the negative set to samples from the same ad topic, is a more meaningful way to evaluate our model.

We see some interesting examples showing that the model understands both the images and statements reasonably well. For example, the result of row 3 in Figure~\ref{fig:psa_ranking}, our ADVISE model ranks safety-related statements higher in the results, moreover, the top-ranked statements all involve the keyword ``helmet'' probably because the model associates the watermelon with head/helmet. Understanding ads is still challenging, and the result of row 4 in Figure~\ref{fig:product_ranking} shows one obstacle. Our model should recognize beer, yet the 'pepsi' mentioned in the second-highest ranked statement is visually quite similar to the bottle in the image thus misleads the model.

% REGION KNN
\clearpage
\section{k-NN retrieval on image regions}
\label{sec:region_knn}

After embedding each image region and weighing them using the attention mechanism, our ADVISE model learns a good image-level representation that could be used to distinguish among statements. Since the image representation is a weighted sum of region representations, regions serve  as visual words in our model and the image is analogous to a sentence.
In order to know if the model also has the ability to assign concept to these visual words (image regions), we made the following qualitative experiments. 

\begin{figure}[t]
    \centering
    \includegraphics[width=1\linewidth]{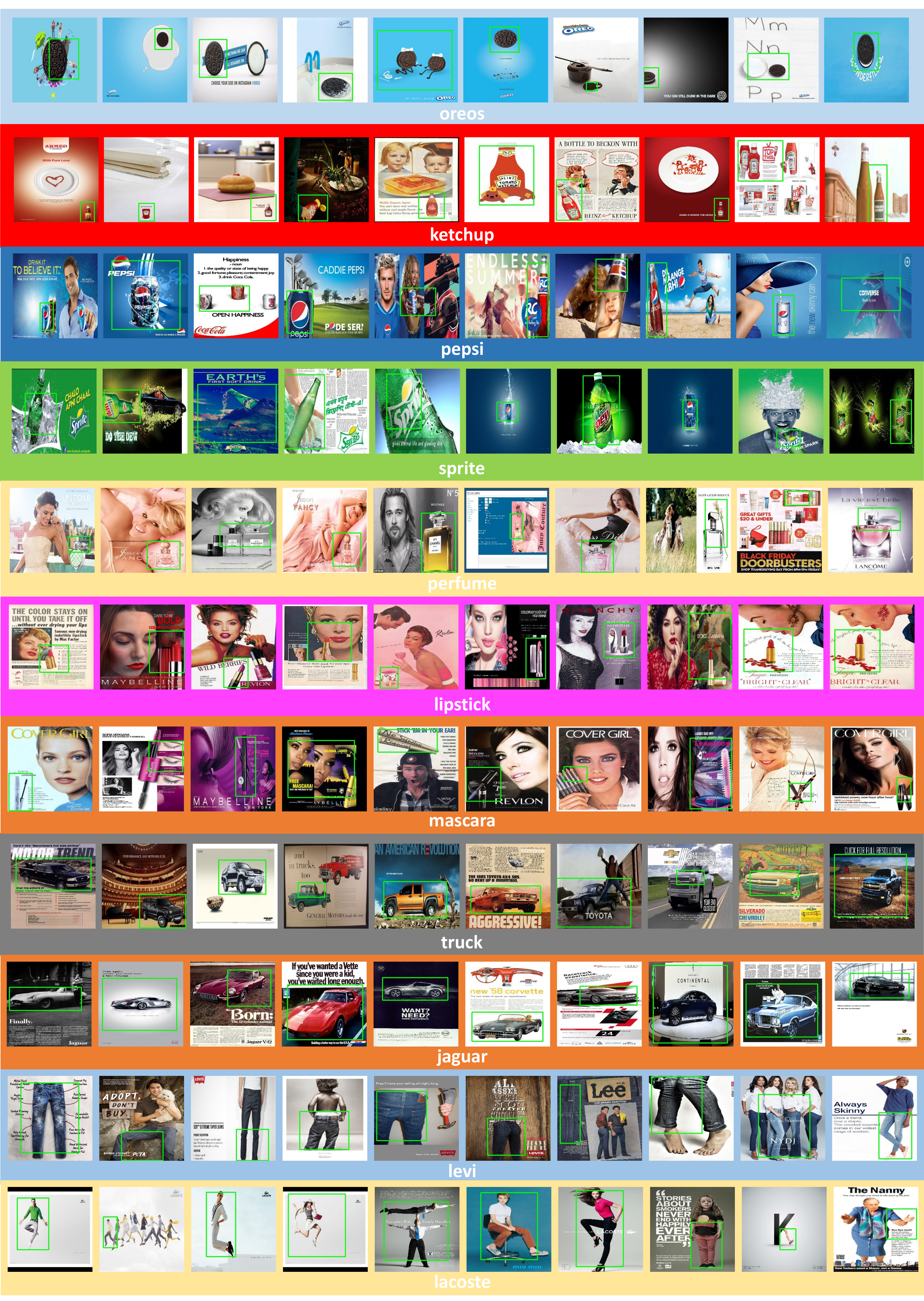}
    \caption{Product words and the retrieved image regions. We select confusing word pairs such as ``pepsi'' and ``sprite'', ``lipstick'' and ``mascara'', ``truck'' and ``jaguar''. We see that our ADVISE model makes mistakes occasionally such as retrieving ``pepsi'' for query ``sprite''. However, the model knows the nuances in general. }
    \label{fig:product_knn}
\end{figure}

\noindent \textbf{Product words as query.}
We choose 11 discriminative words from the vocabulary of product ads and use k-NN to retrieve the 10 most related image regions from the test images. The retrieval results are shown in Figure~\ref{fig:product_knn}. Though we manually choose the 11 words, the k-NN results are entirely generated by the ADVISE model. Please note that we have no such labels associated with regions at training time, that is, we only use the label-agnostic bounding box location annotations (we ignore the symbol categorical labels and semantic meaning of the box region), and the image-level statement annotations, to train the model. However, the model itself successfully learns to associate the concept with a specific region.

\begin{figure}[t]
    \centering
    \includegraphics[width=1\linewidth]{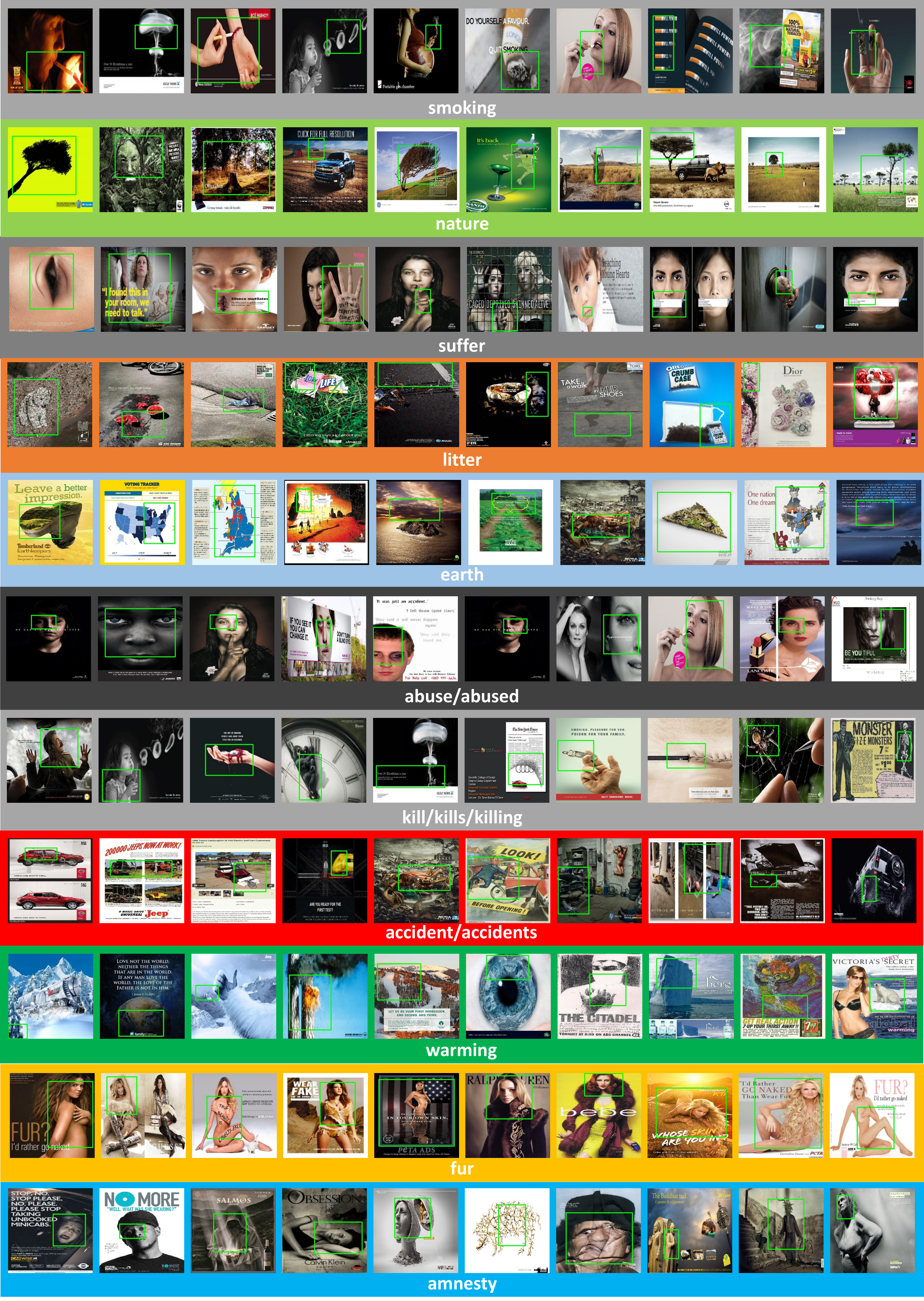}
    \caption{PSA words and the retrieved image regions. It is a more challenging task to associate PSA words with image regions since the words in PSAs tend to be more abstract than that in product ads. The qualitative examples such as ``warming'' and ``litter'' remind us that the embedding of image region and the attention mechanism may also depend on the other regions in the same image (e.g. in case that a beautiful woman exists, polar bear does not symbolize ``warming'' any longer), which is an interesting research direction for our future work.}
    \label{fig:psa_knn}
\end{figure}

\noindent \textbf{PSAs words as query.}
Aligning abstract words from PSA ads to the image regions is a more challenging task. We show in our paper quantitatively that our ADVISE model performs worse in PSAs than that in product ads. To understand the challenge, we show several qualitative examples in Figure~\ref{fig:psa_knn}. Similar to the previous visualization of product ads, we choose multiple discriminative words and retrieve image regions using k-NN. We retrieve the top-20 image regions for each query, merge some queries (such as ``kill'', ``kills'', and ``killing''), and manually select (since the results are not as good as that of product ads) 10 typical examples to visualize.

